\documentclass[journal]{IEEEtran}
\usepackage{algorithm}
\usepackage{algpseudocode}
\usepackage{amssymb,amsmath,amsfonts}
\usepackage{multirow}
\usepackage{booktabs}
\usepackage{textcomp}
\usepackage{ragged2e}
\usepackage{url}
\usepackage{verbatim}
\usepackage{stfloats}
\usepackage{caption}
\usepackage{subcaption}
\usepackage{threeparttable}
\usepackage{cite}
\usepackage[pdftex]{graphicx}
\usepackage{xcolor}
\usepackage{hyperref}
\usepackage{pgfplots}
\usepackage{pgfplotstable}
\def\BibTeX{{\rm B\kern-.05em{\sc i\kern-.025em b}\kern-.08em
    T\kern-.1667em\lower.7ex\hbox{E}\kern-.125emX}}
\usepackage{balance}

\usepackage{forest}
\usetikzlibrary{arrows.meta,shapes,positioning,shadows,trees,calc,matrix}
\usepgfplotslibrary{groupplots,polar}
\pgfplotsset{compat=newest}

\definecolor{mutedRed}{HTML}{C62E2E}
\newcommand{\rev}[1]{#1}
\newcommand{\enlargeX}{3.5}
\newcommand{\barWidth}{9.5pt}
\newcommand{\radarScale}{0.23}
\newcommand{\groupSize}{2 by 3}

\title{
UniTE: A Survey and Unified Pipeline for Pre-training Spatiotemporal Trajectory Embeddings
}

\author{
Yan~Lin,
Zeyu~Zhou, 
Yicheng~Liu, 
Haochen~Lv, 
Haomin~Wen,
Tianyi~Li,
Yushuai~Li, \\
Christian S. Jensen,~\IEEEmembership{Fellow,~IEEE},
Shengnan~Guo,
Youfang Lin,
Huaiyu~Wan

\thanks{Corresponding author: Huaiyu~Wan.}
\thanks{Yan Lin, Zeyu Zhou, Yicheng Liu, Haochen Lv, Haomin Wen, Shengnan Guo, Youfang Lin, and Huaiyu Wan are with the Beijing Key Laboratory of Traffic Data Analysis and Mining, School of Computer Science and Technology, Beijing Jiaotong University, Beijing 100044, China, and the Key Laboratory of Intelligent Passenger Service of Civil Aviation (CAAC), Beijing 101318, China.
Tianyi Li, Yushuai Li, and Christian S. Jensen are with the Department of Computer Science, Aalborg University, Aalborg, 9220, Denmark.}
\thanks{
E-mail:\{ylincs, zeyuzhou, liuyichen, haochenlv, wenhaomin\} @bjtu.edu.cn;
\{tianyi, yusli, csj\}@cs.aau.dk;
\{guoshn, yflin, hywan\}@bjtu.edu.cn.
}
}

\newtheorem{definition}{Definition}

\begin{document}

\maketitle

\begin{abstract}
Spatiotemporal trajectories are sequences of timestamped locations, which enable a variety of analyses that in turn enable important real-world applications. It is common to map trajectories to vectors, called embeddings, before subsequent analyses. Thus, the qualities of embeddings are very important. Methods for pre-training embeddings, which leverage unlabeled trajectories for training universal embeddings, have shown promising applicability across different tasks, thus attracting considerable interest. However, research progress on this topic faces two key challenges: a lack of a comprehensive overview of existing methods, resulting in several related methods not being well-recognized, and the absence of a unified pipeline, complicating the development of new methods and the analysis of methods.

We present UniTE, a survey and a unified pipeline for this domain. In doing so, we present a comprehensive list of existing methods for pre-training trajectory embeddings, which includes methods that either explicitly or implicitly employ pre-training techniques. Further, we present a unified and modular pipeline with publicly available underlying code, simplifying the process of constructing and evaluating methods for pre-training trajectory embeddings. Additionally, we contribute a selection of experimental results using the proposed pipeline on real-world datasets.
\rev{Implementation of the pipeline is publicly available at \url{https://github.com/Logan-Lin/UniTE}.}
\end{abstract}


\IEEEdisplaynontitleabstractindextext
\IEEEpeerreviewmaketitle

\section{Introduction}\label{sec:introduction}
\begin{figure}[t]
    \centering
    \begin{subfigure}[b]{.45\linewidth}
        \includegraphics[width=\linewidth]{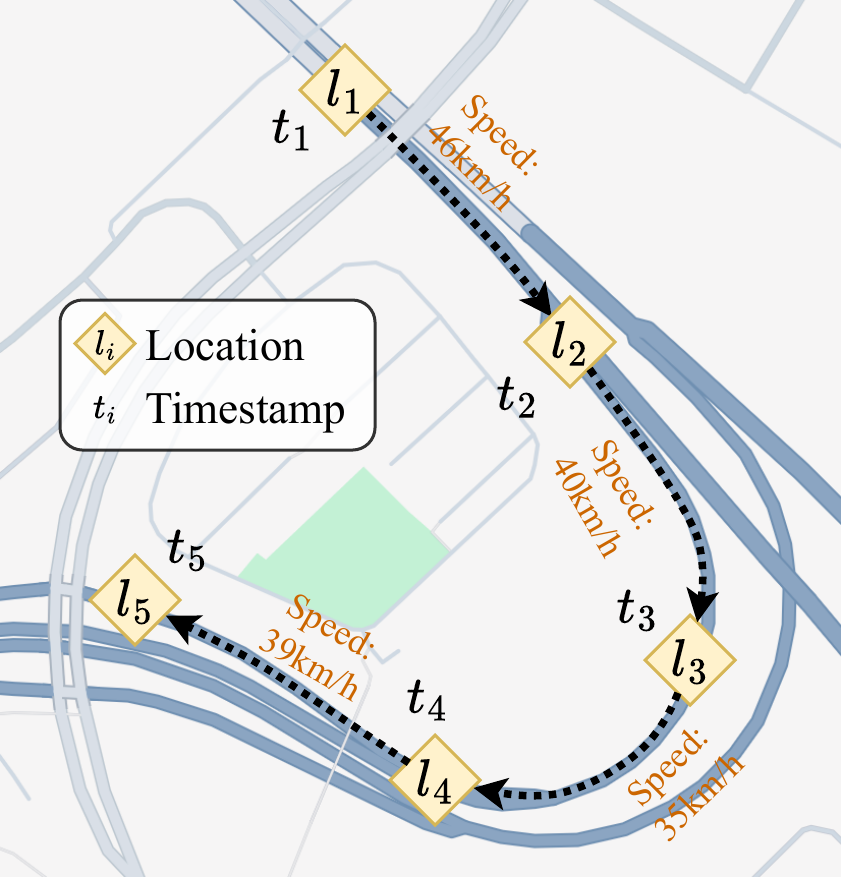}
        \caption{\rev{A vehicle trajectory.}}
        \label{fig:vehicle-trajectory}
    \end{subfigure}
    \hfill
    \begin{subfigure}[b]{.5\linewidth}
        \includegraphics[width=\linewidth]{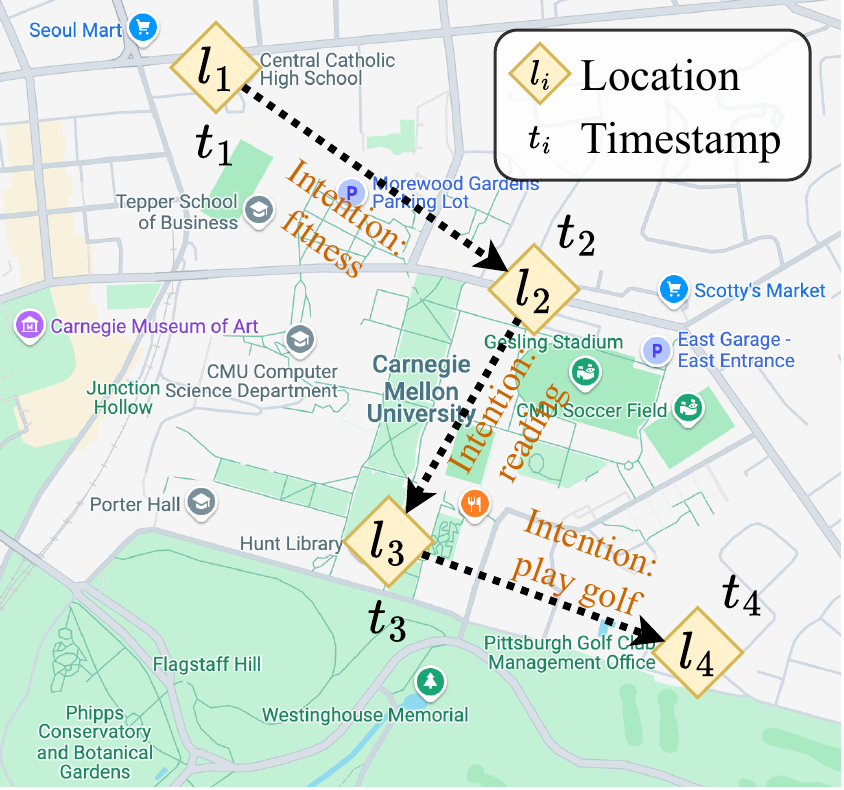}
        \caption{\rev{An individual trajectory.}}
        \label{fig:individual-trajectory}
    \end{subfigure}
    \caption{\rev{Examples of two types of trajectories.}}
    \label{fig:spatiotemporal-trajectories}
\end{figure}

\IEEEPARstart {A} 
spatiotemporal (ST) trajectory is \rev{a sequence of spatial locations associated with specific timestamps} that captures the movement of an individual or an object. \rev{Each trajectory point is a (location, timestamp) pair sampled from the movement.}
\rev{
Figure~\ref{fig:spatiotemporal-trajectories} presents examples of two trajectory types. The vehicle trajectory shown in Figure~\ref{fig:vehicle-trajectory} captures a vehicle's movement, providing insights into driving behavior, route choices, and travel speeds on the road segments it covers. Similarly, the individual trajectory depicted in Figure~\ref{fig:individual-trajectory} illustrates a person's visits to various locations, offering an understanding of their travel intentions and preferences. The wealth of information contained within these trajectories supports the development of numerous tasks in trajectory data analysis and management. These tasks include traffic forecasting~\cite{wang2015predictability,DBLP:conf/aaai/GuoLFSW19,DBLP:conf/aaai/SongLGW20,wang2021libcity}, trajectory classification~\cite{liang2021modeling,zhang2022multi,DBLP:conf/ijcai/LuoZ0WZRL24}, routing~\cite{DBLP:conf/gis/ChondrogiannisB15,DBLP:conf/icde/LiuJYZ18a}, trajectory-based prediction~\cite{DBLP:conf/ijcai/WuCSZW17,kong2018hst,wang2019empowering,wang2021deep,wang2022personalized,liu2024icde}, urban management~\cite{wang2021deepb,wang2021dgeye}, and anomaly detection~\cite{liu2020online,han2022deeptea}.
}

\rev{
The efficient utilization of trajectory data in downstream tasks increasingly relies on machine learning for automating information extraction and executing related activities. In particular, the emergence of deep learning techniques has significantly contributed to their widespread application in trajectory modeling. RNN-based approaches~\cite{wang2019empowering,ji2020interpretable,ji2022stden,wang2023traffic} have been introduced to extract sequential information from trajectories. Similarly, CNN-based and Transformer-based methods~\cite{wang2016traffic,ji2022precision,ji2023spatiotemporal,jiang2023pdformer} are employed to enhance the performance of sequential modeling. GNN-based solutions~\cite{jiang2023continuous,jiang2023selfsupervised,wu2020learning,ji2023spatiotemporal,wu2019learning,wu2020learning} are explored for incorporating spatial dependencies within trajectories. Additionally, novel theories like causal learning~\cite{DBLP:conf/icml/ScholkopfJPSZM12} and maximum entropy~\cite{max-entropy-pri} are incorporated to enhance the performance of trajectory learning~\cite{lin2023pre,DBLP:conf/ijcai/LuoZ0WZRL24}.
}

\rev{
A crucial element in these deep learning models is the embedding vectors of trajectories—$d$-dimensional latent vectors that represent trajectories. These embeddings transform complex trajectory data into a more manageable fixed-size format. They effectively capture essential information from the trajectories, enabling subsequent models to process and learn from them more efficiently. Therefore, the quality and comprehensiveness of these embeddings are pivotal to the performance of deep learning models. The fundamental question remains: how can we obtain effective trajectory embeddings?
}

\begin{figure}
    \centering
    \begin{subfigure}[b]{0.48\linewidth}
        \includegraphics[width=\linewidth]{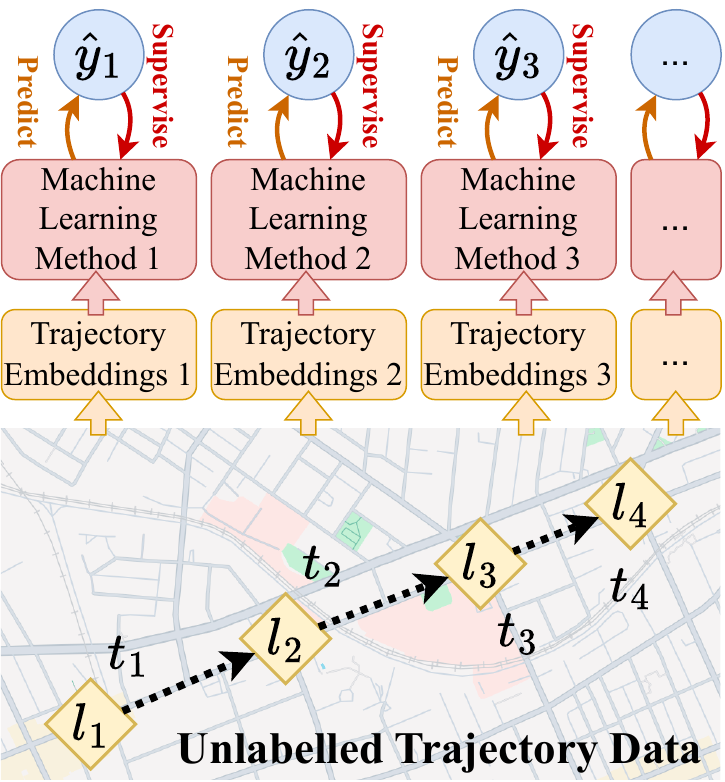}
        \caption{\rev{End-to-end training.}}
        \label{fig:end-to-end}
    \end{subfigure}
    \hfill
    \begin{subfigure}[b]{0.48\linewidth}
        \includegraphics[width=\linewidth]{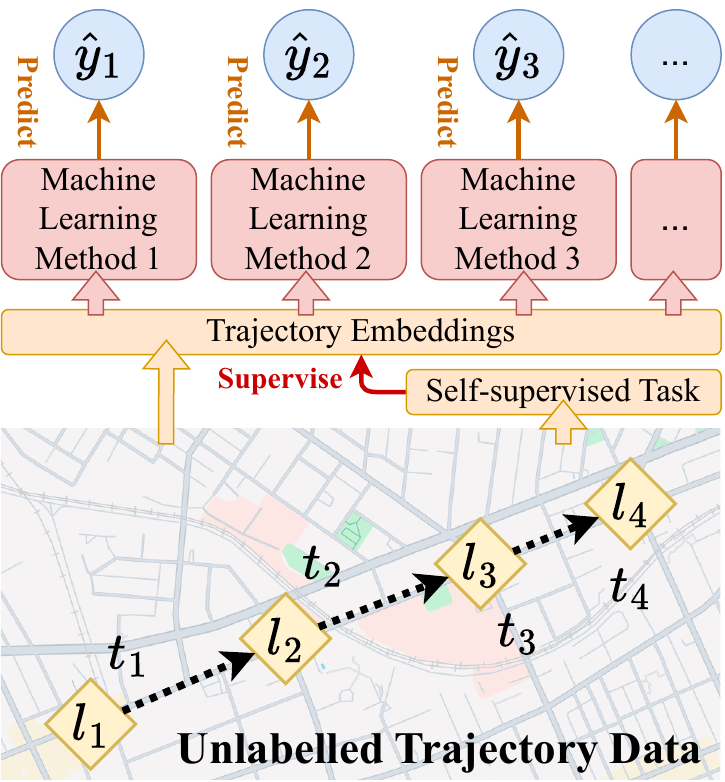}
        \caption{\rev{Pre-training.}}
        \label{fig:pre-training}
    \end{subfigure}
    \caption{\rev{Two approaches to train trajectory embeddings.}}
    \label{fig:training-trajectory-embeddings}
\end{figure}

\rev{
Trajectory embeddings can be trained either end-to-end or through pre-training.
The end-to-end approach~\cite{kong2018hst,DBLP:conf/cikm/LiangOWLCZZZ22,DBLP:journals/www/SangXCZ23} integrates embedding training directly with task-specific supervision, allowing for straightforward implementation within a larger model training pipeline. However, this method typically requires large amounts of labeled trajectory data to perform well. Additionally, models trained end-to-end often have limited transferability across tasks due to their task-specific training objectives, as illustrated in Figure~\ref{fig:end-to-end}.
On the other hand, pre-training embeddings~\cite{tian2020contrastive,chen2020simple,DBLP:journals/tkde/LiuZHMWZT23} offers a promising alternative. This approach involves training trajectory embeddings using task-invariant self-supervised tasks that do not depend on labeled data. By leveraging self-supervised techniques, pre-training can utilize the abundance of available unlabeled trajectory data, thus reducing the need for costly labeled datasets. Moreover, pre-trained embeddings can be reused across multiple downstream tasks without needing retraining from scratch for each new task—significantly enhancing both efficiency and effectiveness, as illustrated in Figure~\ref{fig:pre-training}.
}
Despite the growing interest, research on the pre-training of trajectory embeddings faces two key challenges that if addressed will accelerate advances.

(1)~\textbf{Lack of a comprehensive survey.}
Pre-training is adopted widely for trajectory representation learning. While some methods are designed explicitly for learning universal trajectory embeddings that are shared across various downstream tasks~\cite{lin2023pre}, many methods that focus on one specific downstream task also implicitly employ pre-training techniques for learning trajectory embeddings. For example, some methods use an auto-encoding pre-training framework to measure trajectory similarity and determine distances between trajectories within the embedding space~\cite{li2018deep}. However, such implicit methods are not widely recognized as pre-training methods. Consequently, the potential application of their result embeddings across different downstream tasks remains underexplored. Existing surveys and research on the pre-training of trajectory embeddings tend to focus on methods explicitly designed for learning universal embeddings, thus not considering the full scope of relevant methods.

(2)~\textbf{Lack of a unified pipeline.}
Methods for the pre-training of trajectory embeddings are varied and often described and implemented using disparate frameworks. For example, methods based on contrastive learning~\cite{chen2020simple} initiate the process by augmenting trajectories into categories of targets and positive and negative samples. These samples are subsequently transformed into latent embeddings by one or several encoders. The pre-training employs a contrastive loss function on these embeddings. In contrast, auto-encoding methods~\cite{hinton2006reducing} initially utilize an encoder-decoder pair, followed by pre-training that applies a reconstruction loss to the output of the decoder.
Moreover, the assessment of the effectiveness of embeddings is carried out on diverse tasks and under different experimental settings. Notably, most implicit methods for the pre-training of trajectory embeddings are evaluated using a single downstream task. These variations in methodologies and evaluation criteria complicates the analysis of existing approaches and represent a barrier to the straightforward development and implementation of new methods.

To address these challenges and accelerate research on the pre-training of trajectory embeddings, we present a survey and a unified pipeline named \textit{\underline{Uni}fied \underline{T}rajectory \underline{E}mbeddings} (\textbf{UniTE}). Initially, we present an extensive review of current methods used for the pre-training of trajectory embeddings, covering both explicit and implicit approaches. Following this, we present a unified and modular pipeline designed to standardize the implementation of existing methods and streamline the development of new ones. Additionally, the proposed pipeline facilitates use of embeddings in diverse downstream tasks, allowing for straightforward evaluation and comparison of different methods.
The underlying code is publicly available at \url{https://github.com/Logan-Lin/UniTE}.

\textbf{Related Surveys.} Despite the growing interest in the pre-training approach to trajectory representation learning and the increasing number of contributions on the pre-training of trajectory embeddings, there is a lack of surveys on this specific topic.
While some existing surveys~\cite{DBLP:journals/tist/Zheng15,DBLP:journals/csur/WangBCC21} provide a broad introduction to techniques, methodologies, and tasks utilized in trajectory data mining, they offer only limited coverage of the specific subject of pre-training. Recent surveys~\cite{DBLP:journals/csur/LucaBLP23,DBLP:journals/corr/abs-2402-00732,chen2024deep} have delved deeper into the field of trajectory deep learning, but again touch only briefly on the topic of pre-training of trajectory embeddings briefly.
Surveys also exist that focus on specific aspects of trajectory data management or applications, such as trajectory prediction~\cite{DBLP:journals/dase/YuanL21}, trajectory similarity computation~\cite{DBLP:journals/corr/abs-2303-05012}, and travel time estimation~\cite{DBLP:journals/corr/abs-1904-05037}. However, these surveys do not provide a detailed examination of the pre-training of trajectory embeddings.

In summary, our main contributions are as follows:
\begin{itemize}
    \item We enrich the domain of the pre-training of trajectory embeddings by proposing the first comprehensive survey and unified pipeline on this topic.
    \item We conduct an extensive survey spaning existing methods for the pre-training of trajectory embeddings, covering both explicit and implicit approaches.
    \item We introduce a unified and modular pipeline for standardizing the implementation and evaluation of methods for the pre-training of trajectory embeddings.
    \item We release the code and report on extensive experiments on several real-world trajectory datasets to illustrate the utility of UniTE. 
\end{itemize}

Overall, we hope that this survey and pipeline can accelerate research on trajectory pre-training.

\section{Preliminaries}\label{sec:preliminaries}

\subsection{Pre-training of Embeddings}
Methods for the pre-training of embeddings~\cite{hinton2006reducing,tian2020contrastive,DBLP:journals/tkde/LiuZHMWZT23} aims to equip models with prior knowledge of data features before adopting them for specific tasks.

The main objective of pre-training embeddings is to encode input data into dense vectors that represent the underlying features in a more abstract and easily understandable form for downstream machine learning models.
This is achieved by training the model on a pre-text task, such as predicting the next word in a sentence in natural language processing (NLP) tasks~\cite{DBLP:conf/nips/BengioDV00} or recognizing objects in images without labels in computer vision (CV) tasks~\cite{oord2018representation}. This process enables downstream models to develop a generalized understanding of the data, which can greatly enhance its performance on downstream tasks, even when given relatively smaller amounts of task-specific data.

The pre-training of embeddings has gained widespread popularity in different domains, particularly in the NLP and CV domains. In NLP, models like word2vec~\cite{Mikolov2013Efficient} and BERT~\cite{devlin2018bert} utilize pre-trained embeddings to achieve state-of-the-art performance at tasks such as text classification, question answering, and language generation. In CV, techniques such as pre-trained Convolutional Neural Networks (CNNs)~\cite{chen2020simple} are employed for image classification, object detection, and more, by learning from large-scale image datasets like ImageNet.

\subsection{Definitions}

\begin{definition}
[Spatiotemporal Trajectory]
A spatiotemporal trajectory records the movement of an object during a certain time span. Formally, a trajectory is represented as $\mathcal{T}=\langle (l_1,t_1), (l_2,t_2), \dots, (l_N,t_N) \rangle$. Here, $N$ is the length of the trajectory and $l_i=(\mathrm{lng}_i, \mathrm{lat}_i)$ represents the location of the $i$-th point. The timestamp $t_i$ captures the time when the $i$-th point was recorded.
\end{definition}

\begin{definition}
[Trajectory Dataset]
A trajectory dataset $\mathbb{T}$ is a set of trajectories, where each trajectory $\mathcal T\in \mathbb T$ has been collected in a specific geographical region and time frame.
\end{definition}

\begin{definition}
[Trajectory Embedding]
Given a trajectory $\mathcal T$, its embedding is a fixed-length vector $\boldsymbol z_{\mathcal T}\in \mathbb R^d$, where $d$ is the dimensionality of the embedding.
A trajectory encoder $f_\theta$ with learnable parameters $\theta$ is often used to map variable length trajectories to their embeddings, i.e., $f_\theta(\mathcal T) = \boldsymbol z_{\mathcal T}$.
\end{definition}

\begin{definition}
[Road Network]
A road network is modeled as a directed graph $\mathcal G=(\mathcal V, \mathcal E)$, where $\mathcal V$ is a collection of nodes $v_i$ that correspond to either an intersection of road segments or the end of a segment and $\mathcal E$ is a set of edges $s_i\in\mathcal E$ that correspond to a road segment that connects two nodes. An edge $s_i=(v_j,v_k)$ is characterized by its starting and ending nodes.
\end{definition}

\subsection{Problem Statement}
\noindent \textit{\textbf{Pre-training of Trajectory Embeddings.}}
Given a trajectory dataset $\mathbb T$, a method for the pre-training of trajectory embeddings aims to develop a trajectory encoder $f_\theta$ that maps a trajectory $\mathcal T$ to its embedding vector $\boldsymbol z_\mathcal T$. The encoder is optimized using a specific pre-training objective. The optimization process can be formulated as follows:
\begin{equation}
    \theta = {\arg\min}_\theta \sum_{\mathcal T\in \mathbb T} 
    \mathcal L(f_\theta(\mathcal T)),
\end{equation}
where $\mathcal L$ is the pre-training objective, designed to be independent of any specific task. The trained trajectory encoder $f_\theta$ is then applied to downstream tasks, either through fine-tuning or using unsupervised schemes.

\section{Survey on the Pre-training of Trajectory Embeddings}
\label{sec:survey}
We present a comprehensive inventory of existing methods for the pre-training of trajectory embeddings. 
This inventory includes explicit methods, designed to create universal trajectory embeddings for use in different downstream tasks, and implicit methods, crafted for specific downstream tasks while adopting pre-training techniques to acquire task-invariant embeddings.
Considering the wide range of initial applications for these methods, we arrange them based on the pre-training frameworks they employ, as shown in Figure~\ref{fig:methods}. We proceed to provide a concise overview of the features and implementation of these methods.

\begin{figure}
    \centering
    \input{plot/methods}
    \caption{\rev{Overview of existing methods, arranged based on their pre-training frameworks.}}
    \label{fig:methods}
\end{figure}

\subsection{Word2vec-based Methods}
\label{sec:word2vec-based-methods}
\rev{Word2vec~\cite{DBLP:conf/nips/MikolovSCCD13,Mikolov2013Efficient}, a classical language model, uses a two-layer neural network to learn word embeddings based on the distributional hypothesis~\cite{DBLP:journals/corr/Rong14}. It implements this through two architectures: Continuous Bag-of-Words (CBOW)~\cite{Mikolov2013Efficient}, which predicts a target word from its context, and Skip-Gram~\cite{DBLP:conf/nips/MikolovSCCD13}, which predicts the context from the target word.}

\rev{This approach translates effectively to spatiotemporal trajectory analysis, where locations appearing in similar contexts are assumed to serve similar functions~\cite{feng2017poi2vec,DBLP:journals/tkde/WanLGL22}. Several trajectory embedding methods adopt word2vec's framework, primarily focusing on learning embeddings for individual locations. These location embeddings can then be aggregated into trajectory embeddings through techniques like mean pooling across feature dimensions.}

\subsubsection{FVTI}
\label{sec:methods-fvti}
\rev{FVTI~\cite{huang2021data} is an anomaly detection method for vehicle trajectories that leverages word2vec to generate trajectory embeddings for similarity computation.}

\rev{For a trajectory $\mathcal T$, FVTI treats each point $(l_i,t_i)$ as a word in a sentence. The embedding process consists of three phases: First, FVTI extracts three features (range, range rate, and speed) from each point and quantizes them into discrete tokens, effectively transforming each point into a three-token word. Second, it employs the CBOW model of word2vec with a one-word context window to generate embedding vectors for each unique word constructed in the first phase. Third, it obtains the trajectory embedding by averaging the word embedding vectors.}

\subsubsection{GCM}
\label{sec:methods-gcm}
\rev{GCM~\cite{DBLP:journals/access/WangFSALWHC20} is an vehicle trajectory anomaly detection method that maps trajectories to embedding vectors for binary classification. Its trajectory embedding component utilizes word2vec as its foundation.}

\rev{GCM represents each trajectory point $(l_i,t_i)$ as a word by partitioning the geographical region into a uniform grid and associating each cell with an embedding vector, with each point being associated with the embedding vector of the cell it belongs to. Using the Skip-Gram architecture of word2vec, GCM trains these embeddings by optimizing the prediction of middle points given the trajectory's start point and end point.}

\subsubsection{POI2Vec}
\label{sec:methods-poi2vec}
\rev{POI2Vec~\cite{feng2017poi2vec} learns location representations from individual trajectories using word2vec's framework to generate embedding vectors for locations that can be used in location trajectory analysis.}

\rev{POI2Vec adapts the CBOW model by incorporating spatial correlations through a hierarchical division of the geographical region into cells. Each cell maps to a node in a binary tree used for hierarchical softmax calculations, with locations assigned based on their containing cell. The embedding vectors are then trained using word2vec on the trajectory data.}

\subsubsection{TALE}
\label{sec:methods-tale}
\rev{TALE~\cite{wan2019learning,DBLP:journals/tkde/WanLGL22} is a word2vec-based method that learns location embeddings from individual trajectories for trajectory prediction and location classification tasks.}

\rev{TALE extends CBOW by incorporating temporal correlations through a hierarchical tree structure. It partitions each day into time spans represented as nodes in a multi-branch tree, with each node containing a Huffman sub-tree based on location visit frequencies. This structure enables temporal-aware hierarchical softmax calculations during the word2vec training process.}

\subsection{Masked Language Model-based Methods}
\label{sec:masked-language-model-based-methods}
\rev{Masked Language Model (MLM) represent a significant advancement in self-supervised learning and natural language processing, particularly through BERT~\cite{devlin2018bert}.
MLM operates by masking random tokens in a sequence with a special mask token and training the model to predict these masked tokens using the surrounding context. This approach enables the model to learn rich contextual representations that capture both local and broader linguistic patterns.
BERT additionally employs Next Sentence Prediction (NSP), where the model learns to determine if two sentences are sequential in the original text. NSP helps develop understanding of inter-sentence relationships and textual coherence.}

\rev{The success of MLM and NSP has led to their adoption beyond NLP, including applications in spatiotemporal location and trajectory embedding.}

\subsubsection{CTLE}
\label{methods:ctle}
\rev{CTLE~\cite{lin2021pre} is a multi-task pre-training framework that employs MLM to learn contextual embeddings of locations from individual trajectories for trajectory prediction.}

\rev{Following the MLM paradigm, CTLE randomly masks trajectory points $(l_i,t_i)$ in a trajectory $\mathcal T$ by replacing their location and time features with a mask token. A Transformer-based encoder processes the masked trajectory to generate embeddings for the masked points, which are then used to predict the original features. The final trajectory embedding is obtained by mean pooling the point embeddings produced by the pre-trained encoder.}

\subsubsection{Toast}
\label{methods:toast}
\rev{Toast~\cite{DBLP:conf/cikm/ChenLCBLLCE21} is a BERT-based~\cite{devlin2018bert} method that learns embeddings for both road segments and vehicle trajectories.}

\rev{Toast adapts BERT's MLM and NSP tasks for trajectory pre-training. It first generates preliminary road segment embeddings using word2vec's Skip-Gram model on random walks of the road network $\mathcal G$. A Transformer encoder then processes these embeddings to generate trajectory representations, with trajectories map-matched to sequences of road segments on $\mathcal G$. The model is trained using masked road segment reconstruction and trajectory sequence prediction tasks. The final trajectory embedding is obtained by mean pooling the encoder outputs.}

\subsection{Auto-encoding-based Methods}
\label{sec:auto-encoding-based-methods}
\rev{The Auto-Encoding (AE) framework~\cite{hinton2006reducing} is a fundamental self-supervised learning approach for efficient encoding of unlabeled data. The framework employs an encoder that compresses input data into low-dimensional vectors and a decoder that reconstructs the original data from these vectors. Through this process of compression and reconstruction, the encoder learns to preserve essential features of the input data.}

\rev{The AE framework has been widely adopted for dimensionality reduction~\cite{DBLP:conf/nips/OordVK17}, denoising~\cite{DBLP:journals/jmlr/VincentLLBM10}, anomaly detection~\cite{liu2020online}, and generative modeling~\cite{DBLP:journals/corr/ZhaoSE17b}. It is particularly valuable for learning trajectory embeddings, enabling the conversion of variable-length sequences into fixed-length vectors.}

\subsubsection{DTC}
\label{sec:methods-dtc}
\rev{DTC~\cite{DBLP:journals/tgis/WangLWWX0WWD22} is a deep learning-based clustering method for vehicle and synthetic trajectories that learns trajectory embeddings for $k$-means clustering through a two-phase AE framework.}

\rev{The method first preprocesses trajectories by discretizing the geographical region into a grid and mapping trajectory points to grid cell tokens. It then employs an RNN-based~\cite{DBLP:journals/cogsci/Elman90} encoder-decoder architecture for pre-training, where the encoder generates trajectory embeddings by learning to reconstruct the tokenized trajectories through the decoder.}

\subsubsection{trajectory2vec}
\label{sec:methods-trajectory2vec}
\rev{trajectory2vec~\cite{yao2017trajectory} employs the AE framework with LSTM-based~\cite{hochreiter1997long} encoder and decoder to derive embeddings for vessel trajectories.}

\rev{The method prepares trajectories by extracting moving behavior sequences using a sliding window. The window moves across each trajectory with half-width steps, extracting features like time intervals, position changes, speed, and turn rates. These behavioral features are then processed through the AE framework to learn the trajectory's embedding vector.}

\subsubsection{TremBR}
\label{sec:methods-trembr}
\rev{TremBR~\cite{fu2020trembr} employs the AE framework to pre-train vehicle trajectory embeddings for various downstream tasks.}

\rev{The method consists of three stages: map-matching trajectories to a road network $\mathcal G$, learning road segment embeddings using word2vec's CBOW model, and generating trajectory embeddings. After map-matching associates each trajectory point with a road segment and timestamp, the road segments are embedded using CBOW. These segment embeddings, combined with timestamps, represent trajectory points. An encoder then processes sequences of these point representations to generate trajectory embeddings, with pre-training guided by trajectory reconstruction.}

\subsubsection{CAETSC}
\label{sec:methods-caetsc}
\rev{CAETSC~\cite{liang2021unsupervised} is a deep learning-based method that computes vessel trajectory similarity by mapping trajectories to embedding vectors.}

\rev{CAETSC operates in two phases: First, it converts trajectories into image representations by partitioning the geographical region into a uniform grid, where each pixel indicates trajectory presence in the corresponding grid cell. Second, it employs a CNN-based~\cite{DBLP:conf/nips/CunBDHHHJ89} encoder-decoder architecture within the AE framework to transform these images into low-dimensional trajectory embeddings.}

\subsection{Variational Auto-encoding-based Methods}
\rev{The Variational Auto-Encoder (VAE) framework~\cite{kingma2013auto} extends the vanilla AE framework by incorporating variational Bayesian methods. Rather than encoding inputs as fixed embeddings, VAE models them as distributions in the embedding space - specifically, the encoder outputs parameters of a multivariate Gaussian distribution from which embeddings are sampled for reconstruction. This probabilistic approach enables both data generation and more meaningful embeddings.}

\rev{The VAE framework's effectiveness has been demonstrated across domains like synthetic data generation~\cite{DBLP:conf/nips/RazaviOV19} and distribution modeling~\cite{DBLP:conf/nips/OordVK17}, including several methods for learning robust trajectory embeddings.}

\subsubsection{GM-VSAE}
\rev{GM-VSAE~\cite{liu2020online} employs the VAE framework to detect anomalous vehicle trajectories by modeling trajectory data as Gaussian distributions in the embedding space. By comparing original and generated trajectories, it enables anomaly detection through the VAE's generative capabilities.}

\rev{The method uses LSTM-based~\cite{hochreiter1997long} encoder and decoder networks. For a trajectory $\mathcal T$, the encoder maps it to parameters of a multivariate Gaussian distribution. The decoder samples from this distribution to reconstruct the trajectory's locations, with the model optimized using reconstruction and VAE regularization losses.}

\subsubsection{TrajODE}
\rev{TrajODE~\cite{liang2021modeling} leverages the VAE framework to learn versatile vehicle trajectory embeddings for multiple downstream applications.}

\rev{TrajODE introduces a novel approach using Neural Ordinary Differential Equations (NeuralODEs)~\cite{hartman2002ordinary,DBLP:conf/nips/ChenRBD18} in its encoder-decoder architecture. The encoder processes trajectories through an LSTM network whose hidden states are updated via an ODE solver and spatiotemporal gating mechanism, mapping each trajectory $\mathcal T$ to a multivariate Gaussian distribution. The embedding is refined using a Continuous Normalizing Flow (CNF)~\cite{DBLP:conf/nips/ChenRBD18} before the decoder samples from it to reconstruct the original trajectory.}

\subsection{Denoising Auto-encoding-based Methods}
\rev{The denoising auto-encoder (DAE)~\cite{DBLP:conf/icml/VincentLBM08} framework extends the AE framework by incorporating noise robustness. While AE simply reconstructs inputs from compressed representations, DAE intentionally corrupts input data and tasks the decoder with recovering the original, uncorrupted version. This approach yields more robust feature extraction and representation learning, particularly valuable for denoising tasks.}

\rev{The DAE framework has proven effective for pre-training trajectory embeddings, especially when working with noisy or sparse trajectory data.}

\subsubsection{t2vec}
\label{sec:methods-t2vec}
\rev{The t2vec~\cite{li2018deep} method employs the DAE framework to learn robust vehicle trajectory embeddings for similarity computation, particularly effective for noisy trajectories.}

\rev{The method operates in two stages. First, it preprocesses trajectories by randomly dropping points and adding Gaussian noise to remaining points $(l_i,t_i)$ in trajectory $\mathcal T$, then maps points to discrete tokens using a uniform grid. Second, it uses RNN-based~\cite{DBLP:journals/cogsci/Elman90} encoder-decoder architecture to learn trajectory embeddings, with the encoder compressing trajectories and the decoder reconstructing them. The model is pre-trained using a spatial proximity-aware loss function.}

\subsubsection{Robust DAA}
\label{sec:methods-robustdaa}
\rev{Robust DAA~\cite{DBLP:conf/mdm/ZhangXJX0L19} employs the DAE framework to learn individual and flight trajectory embeddings for clustering tasks.}

\rev{The method first converts trajectories into image representations encoding movement patterns. Using an attention-based~\cite{vaswani2017attention} encoder-decoder architecture, it decomposes each trajectory image into modelable ($\mathcal T_D$) and noisy ($\mathcal T_N$) components. The model aims to accurately reconstruct $\mathcal T_D$ while suppressing $\mathcal T_N$.}

\subsubsection{TrajectorySim}
\label{sec:methods-trajectorysim}
\rev{TrajectorySim~\cite{DBLP:journals/www/ChenLZCS23} employs the DAE framework for vehicle trajectory similarity computation.}

\rev{The method enhances robustness by applying dropout and Gaussian noise to trajectories. It discretizes points using uniform spatial and temporal grids, then leverages an RNN-based encoder-decoder architecture to learn trajectory embeddings, optimized with a spatiotemporal proximity-aware loss.}

\subsection{Contrastive learning-based Methods}
\rev{Contrastive learning~\cite{chen2020simple,tian2020contrastive} is a self-supervised learning approach that learns embeddings by distinguishing between similar (positive) and dissimilar (negative) data point pairs. Through this discriminative process, the model learns to encode meaningful representations of the data.}

\rev{Having demonstrated success in computer vision and natural language processing~\cite{oord2018representation,tian2020contrastive,chen2020simple}, this approach has also shown promise in trajectory embedding learning.}

\subsubsection{PreCLN}
\rev{PreCLN~\cite{yan2022precln} employs contrastive learning for pre-training vehicle trajectory embeddings to enhance trajectory prediction.}

\rev{The method consists of two key components: trajectory augmentation and dual-view contrastive learning. The augmentation module generates grid view and map-matched view representations for each trajectory $\mathcal T$ - the former discretizes the trajectory into tokens while the latter applies map-matching. These views are further enriched through multi-hop sampling and segmentation. A Transformer-based encoder~\cite{vaswani2017attention} then processes these views into embeddings, with the model trained to differentiate between matching views of the same trajectory versus views from different trajectories.}

\subsubsection{TrajCL}
\label{sec:methods-trajcl}
\rev{TrajCL~\cite{DBLP:conf/icde/Chang0LT23} employs contrastive learning to pre-train vehicle trajectory embeddings for similarity computation.}

\rev{The method generates two views of each trajectory through a three-step process: (1) trajectory augmentation via point shifting, masking, truncation, and simplification; (2) feature enhancement by mapping locations to grid cells and computing spatial angles between consecutive points; and (3) trajectory encoding using self-attention to obtain embeddings. These embeddings are then pre-trained using contrastive learning to maximize the similarity between views of the same trajectory.}

\subsubsection{START}
\rev{START~\cite{DBLP:journals/corr/abs-2211-09510} combines contrastive learning and MLM to learn versatile vehicle trajectory embeddings.}

\rev{The method first map-matches trajectories to the road network and applies augmentations like trimming, masking, and feature corruption. A Transformer encoder~\cite{vaswani2017attention} maps augmented trajectories to embeddings, which are trained to distinguish between augmentations of the same versus different trajectories. The model is jointly optimized using contrastive learning and MLM losses, where MLM predicts masked trajectory points.}

\begin{figure*}[t]
    \centering
    \includegraphics[width=1.0\linewidth]{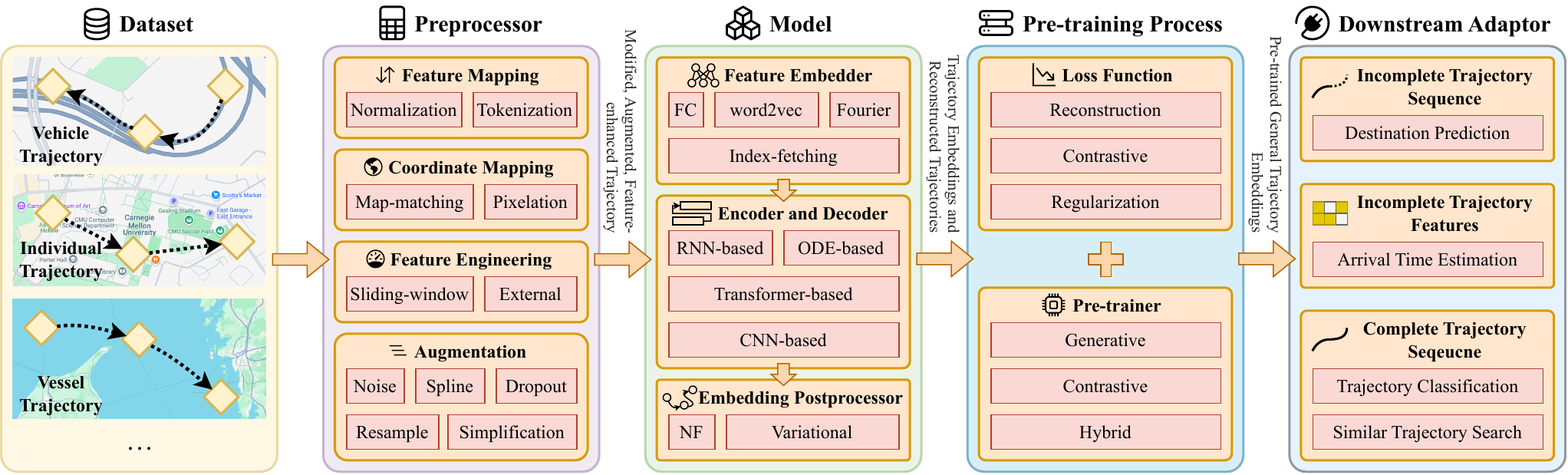}
    \caption{\rev{The UniTE pipeline.}}
    \label{fig:pipeline}
\end{figure*}

\subsubsection{LightPath}
\label{sec:methods-lightpath}
\rev{LightPath~\cite{DBLP:conf/kdd/YangHGYJ23} combines contrastive learning and denoising auto-encoding for pre-training path or map-matched vehicle trajectory embeddings.}

\rev{The method first map-matches trajectories to road segments and randomly removes segments to improve efficiency. It then employs a dual-encoder architecture - a main and auxiliary Transformer encoder - to generate embeddings from two views with different drop ratios for contrastive learning. Additionally, a Transformer decoder reconstructs the original trajectory from the reduced version through denoising auto-encoding.}

\subsubsection{MMTEC}
\label{sec:methods-mmtec}
\rev{MMTEC~\cite{lin2023pre} leverages information entropy theory to enhance contrastive learning for general vehicle trajectory embeddings applicable across diverse tasks.}

\rev{The method generates dual trajectory views: a travel semantics view via an attention-based encoder on map-matched data, and a continuous spatiotemporal view using Neural Controlled Differential Equations (CDE)~\cite{kidger2020neural}. Through Maximum Multi-view Trajectory Entropy Coding, MMTEC optimizes information entropy while maintaining view consistency.}

\begin{table*}
\centering
\caption{Information and statistics of spatiotemporal trajectory datasets.}
\resizebox{1.0\linewidth}{!}{
\begin{tabular}{c|ccccccc}
  \toprule

   Dataset & Source & Time Span & Longitude Scope & Latitude Scope & \# Unique Tokens & \# Trajectories & \# Points \\
  \midrule
  Chengdu & Taxi & 10.01  --  11.30, 2018 & 104.04300  --  104.12654 & 30.65523  --  30.72699 & 2,505 & 121,394 & 3,032,212 \\
  Xian & Taxi & 10.01 -- 11.30, 2018 & 108.9219  --  109.0088 & 34.20495  --  34.27860 & 3392 & 202,849 & 18,267,443 \\
  Porto & Taxi & 07.01 -- 11.01, 2013 & -8.651997  --  -8.578008 & 41.14201  --  41.17399 & 2,225 & 1,095,821 & 12,852,750 \\
  T-Drive~\cite{DBLP:conf/gis/YuanZZXXSH10} & Taxi & 02.02 -- 02.08, 2008 & Beijing & Beijing & - & 2,100,000 & 17,600,000 \\
  LaDe \cite{wen2023lade} & Delivery & 6 months &  5 cities & 5 cities & 343,368 & 619,000 & 10,677,000 \\
  \midrule
  Foursquare-TKY & Check-in & 04.04, 2012 -- 02.16, 2013 & Tokyo & Tokyo & 61,857 & - & 537,703 \\
  Foursquare-NYC & Check-in & 04.03, 2012 -- 02.15, 2013 & New York & New York & 38,332 & - & 227,428 \\
  Gowalla & Check-in & 11, 2010 -- 12, 2011 & Global & Global & 1,280,969 & - & 6,442,892 \\
  GeoLife & Mobile Device & 04, 2017 -- 10, 2011 & Beijing & Beijing & - & 17,621 & 23,667,828 \\
  \midrule
  CFD~\cite{liang2021unsupervised} & Vessel & 01.01 -- 01.30, 2018 & 118.291206  --  118.558596 & 38.772082  --  38.897793 & - & 1391 & - \\
  CSC~\cite{liang2021unsupervised} & Vessel & 01.01 -- 01.30, 2018 & 122.761476  --  123.162561 & 37.523868  --  37.715612 & - & 1047 & - \\
  YRE & Vessel & 01.08 -- 08.30, 2017 & 121.437999  --  121.975002 & 31.262001  --  31.531999 & - & 1397 & - \\

  \bottomrule
\end{tabular}
}
\label{table:datasets}
\end{table*}

\section{A Unified and Modular Pipeline}
\label{sec:pipeline}
To standardize the implementation and evaluation of methods for the pre-training of trajectory embeddings, we propose the UniTE pipeline. This pipeline modularizes pre-training methods into five key types of components: dataset, preprocessor, model, pre-training process, and downstream adaptor, as shown in Figure~\ref{fig:pipeline}. We provide a detailed presentation of these components, which can be combined to implement the methods presented in Section~\ref{sec:survey}.

\subsection{Dataset}
\label{sec:dataset}
The dataset component is the core component of UniTE, offering real-world spatiotemporal trajectories for analysis, embedding pre-training, and evaluation. Each dataset $\mathbb{T}$ of trajectories is accompanied by contextual information such as the road network $\mathcal{G}$ that covers the region covered by the trajectories in $\mathbb{T}$.
Table~\ref{table:datasets} lists trajectory datasets used frequently in studies of pre-training methods along with information and statistics.

\rev{\textbf{Vehicle Trajectory Datasets} are typically composed of GPS trajectories of vehicles, such as taxis, buses, and trucks. Here we list five commonly-used vehicle trajectory datasets: Chengdu, Xian, Porto, T-Drive, and LaDe.}
The \textit{Chengdu} and \textit{Xian} datasets, which are released by Didi\footnote{\url{https://gaia.didichuxing.com/}} and include GPS trajectories of taxis operating in Chengdu and Xian, China. The \textit{Porto} dataset, made available on Kaggle\footnote{\url{https://www.kaggle.com/competitions/pkdd-15-predict-taxi-service-trajectory-i/data}} for a taxi trajectory prediction contest, contains GPS trajectories of taxis in Porto, Portugal. \textit{T-Drive}~\cite{DBLP:conf/gis/YuanZZXXSH10}, published by Microsoft, comprises GPS trajectories of taxis in Beijing, China. \textit{LaDe}~\cite{wen2023lade} is released by Cainiao and includes trajectories from last-mile deliveries in five cities.

\rev{\textbf{Individual Trajectory Datasets} are composed of trajectories of individuals' movements. Individual trajectories tend to have richer semantic information regarding the activities and preferences of individuals. Here we list four commonly-used individual trajectory datasets: Foursquare-TKY, Foursquare-NYC, Gowalla, and Geolife.}
\textit{Foursquare-TKY} and \textit{Foursquare-NYC}, released by Foursquare\footnote{\url{https://sites.google.com/site/yangdingqi/home/foursquare-dataset}}, are check-in trajectory datasets recording visits to locations in Tokyo, Japan, and New York City, USA, respectively. \textit{Gowalla}, another check-in trajectory dataset released by Gowalla\footnote{\url{https://www.kaggle.com/datasets/bqlearner/gowalla-checkins}}, covers global check-in records. \textit{Geolife}, a trajectory dataset collected and released by Microsoft\footnote{\url{https://www.microsoft.com/en-us/research/publication/geolife-gps-trajectory-dataset-user-guide/}}, consists of data recorded by mobile devices of 182 users during their daily activities.

\rev{\textbf{Vessel Trajectory Datasets} are typically composed of trajectories of vessels, such as ships and boats. Compared to vehicle trajectories, vessel trajectories tend to be more sparse and cover larger geographical areas since ships typically travel longer distances over open water. Here we list three commonly-used vessel trajectory datasets: CFD, CSC, and YRE.}
\textit{CFD} and \textit{CSC} are vessel trajectory datasets used in the CAETSC study~\cite{liang2021unsupervised} that record the movement of vessels at sea. \textit{YRE}, released by the University of Rhode Island\footnote{\url{https://www.cs.rit.edu/~rlaz/datasets.html}}, covers global vessel trajectories.

Some datasets include unique tokens that can be utilized by the preprocessor that will be introduced later, such as road segments in taxi datasets or points of interest (POIs) in check-in datasets. Road segment data for taxi datasets can be obtained from publicly available services like OpenStreetMap\footnote{\url{https://www.openstreetmap.org/}}, and POI information for check-in datasets can be retrieved from location-based services like AMap\footnote{\url{https://lbs.amap.com/api/javascript-api-v2}}.

\subsection{Preprocessor}
\label{sec:preprocessor}
\rev{
The preprocessor component in UniTE is tasked with converting raw trajectory data into a structured format that is ready for encoding and decoding. The preprocessor involves four major types of operations: feature mapping, coordinate mapping, feature engineering, and augmentation. The framework implements these through several specific preprocessing operations.
}

\subsubsection{\rev{Feature Mapping}}
\label{sec:feature-mapping}
\rev{These operations transform raw features into standardized ranges or discrete tokens. Specifically, two feature mapping operations are implemented: normalization and tokenization.}

\textbf{Normalization} scale the continuous raw features within trajectories to a uniform range (such as $[0,1]$ or $[-1,1]$), thus solving the problem that the value ranges of different features are too different. If the values of some features are too large and exceed by far the values of other features, the results of model training will be dominated by such features, and useful information contained in features with small values will be missed. The two primary methods employed are min-max normalization and z-score normalization.
For a raw feature $x$, \textit{min-max} normalization first subtracts its minimum value and then divides it by the range of $x$:
\begin{equation}
    x_\mathrm{norm} = (x - x_\mathrm{min}) / (x_\mathrm{max} - x_\mathrm{min}),
\end{equation}
where $x_\mathrm{min}$ and $x_\mathrm{max}$ represent the minimum and maximum values of $x$ respectively, and the normalized result $x_\mathrm{norm}$ is in range $[0, 1]$. The \textit{z-score} normalization makes the mean and standard deviation of the results 0 and 1 respectively, which facilitates rapid training. The formula is as follows:
\begin{equation}
    x_\mathrm{norm} = (x - \bar x) / \mathrm{STD}(x),
\end{equation}
where $\bar x$ is the average value of $x$ and $\mathrm{STD}(x)$ is its standard deviation.

\textbf{Tokenization} converts each point in a trajectory into a discrete token. Considering a 1-dimensional feature $x$, we define a series of buckets as:
\begin{equation}
    \langle (-\infty, b_1), [b_1, b_2), [b_2,b_3), \dots, [b_M, +\infty) \rangle
    \label{eq:tokenize}
\end{equation}
The token for $x$ is determined by the bucket it falls into. 

For multi-dimensional features, each dimension is tokenized independently, upon which tokens are merged to form compound tokens. For instance, a point $l_i$ has its longitude and latitude tokenized separately using Equation~\ref{eq:tokenize}, upon which the tokens are combined into a dual-token representation.

\subsubsection{\rev{Coordinate Mapping}}
\label{sec:coordinate-mapping}
\rev{These operations convert trajectory points into different coordinate systems or representations. Specifically, two coordinate mapping operations are implemented: map-matching and pixelation.}

\textbf{Map-matching} can be viewed as mapping GPS locations on to the road network, a non-Euclidean space. GPS location data in a trajectory often does not align accurately with the road due to a range of technical issues. Therefore, it is beneficial to apply map-matching~\cite{DBLP:conf/vldb/BrakatsoulasPSW05} to GPS data to associate it with the road network. Map-matching involves comparing a trajectory with the underlying road network to find the most likely road-network path of the trajectory, thereby transforming the GPS data into path data. This preprocessor uses map-matching algorithms to align a trajectory with a road network $\mathcal{G}$. Each location $l_i$ in the trajectory is mapped to an index $s_i$, indicating the road segment that is closest to $l_i$. Thus, the GPS location $l_i$ is translated into the road segment $s_i$, allowing the current position of the trajectory to be described using roads and intersections.

\textbf{Pixelation} converts a trajectory into an image, and can be viewed an extended case of the tokenization operation applied to 2-dimensional points. It first partitions the spatial regions using a uniform grid with $W\times H$ cells. These cells are analogous to pixels in an image; thus, each trajectory is depicted as a 3D image $\boldsymbol T$ with dimensionality $W\times H\times C$, where $C$ represents the number of channels. The value of a pixel $\boldsymbol T_{i,j}$ is determined by the characteristics of the trajectory points located in the cell corresponding to the pixel. This includes information such as the presence of the trajectory within cell (indicated by a mask), the timestamp, the speed, and the rate of turn.

\subsubsection{\rev{Feature Engineering}}
\label{sec:feature-engineering}
\rev{These operations generate higher-order or additional features from raw trajectory data. Specifically, two feature engineering operations are implemented: sliding-window and external feature extraction.}

\textbf{Sliding-window} captures segments of a trajectory using a window of predefined length and step size, transforming the raw data into fixed-length samples. For example, a window of duration $\delta$ includes trajectory points as follows:
\begin{equation}
\begin{split}
    \langle &(l_i,t_i), (l_{i+1}, t_{i+1}), \dots, (l_{i+j},t_{i+j}) \rangle s.t.\\
    &t_{i+j}-t_i \leq \delta \wedge t_{i+j+1}-t_i > \delta
\end{split}
\end{equation}

Points in these samples are then aggregated to compute higher-order features such as total distance traveled, time traveled, speed, and turning rate.

\rev{
\textbf{External feature extraction} extracts features from external sources to expand the features beyond raw spatiotemporal features. For example, the points in an individual trajectory can be associated with additional information of POIs, including the category and description of the POIs. The points in a vehicle trajectory can be associated with the level and length of their corresponding road segments.
}

\subsubsection{\rev{Augmentation}}
\label{sec:augmentation}
\rev{
Trajectory representation learning often requires substantial amounts of trajectory data to work well. However, sufficient data is not always readily available. Data augmentation addresses this shortage by artificially increasing the cardinality of a dataset. This process involves generating new data points from existing ones by perturbing the data. On the other hand, augmentation can also be used to create multiple views of a trajectory to support contrastive learning or the training of robust embeddings.
Five types of augmentation operations are implemented in UniTE: noise addition, spline interpolation, dropout, resampling, and simplification.
}

\rev{\textbf{Noise addition} involves adding noise to trajectory features to mimic the uncertainty of traffic conditions or the inaccuracy of trajectory recording devices.} Specifically, a feature $x$ is augmented by introducing Gaussian noise with standard deviation $\sigma$:
\begin{equation}
    x' = x + \delta_x, \delta_x \sim \mathcal{N}(0, \sigma),
\end{equation}
where $\sigma$ controls the magnitude of the noise.

\rev{\textbf{Spline interpolation} is designed to reconstruct the continuous dynamics of trajectories using spline functions~\cite{DBLP:journals/jacm/Ferguson64}.} A spline function consists of segmented polynomials defined over a subspace, adhering to certain continuity conditions. Given $n+1$ timesteps $t_0, t_1, \ldots, t_n$, where $t_0 < t_1 < \ldots < t_n$, these steps are referred to as knots. For a specified integer $k \geq 0$, a k-order spline function with knots $t_0, t_1, \ldots, t_n$ is a function $S$ that meets two criteria: (1) $S$ is a polynomial of order $n \leq k$ in each interval $[t_{i-1}, t_i]$, and (2) $S$ has a continuous derivative of order $k-1$ over $[t_1, t_n]$. Specifically, the preprocessor converts a given trajectory $\mathcal{T}$ into a cubic Hermite spline $\boldsymbol{T}$, ensuring that $\boldsymbol{T}_{t_i} = l_i$ for each trajectory point $(l_i, t_i)$.

\rev{
\textbf{Dropout} entails randomly omitting points or spatiotemporal features from a trajectory. For \textit{point dropout}, each point $(l_i, t_i)$ in a trajectory can be randomly deleted with probability $p \in [0,1)$, where $p$ is a hyperparameter to control the drop ratio.
}

\rev{Another way to perform dropout is \textit{feature dropout}, where trajectory features are randomly substituted with a mask token or zeros.}
A trajectory feature $x$ can be masked or zeroed with probability $p \in [0,1)$. A masking function $\mathrm{mask}(\cdot)$ can generate the replaced trajectory features $x_\mathrm{mask}$:
\begin{equation}
    x_\mathrm{mask} = \mathrm{mask}(x) = 
    \left\{
        \begin{aligned}
         &x && \mathrm{if}\ k > p \\
         &[m]\ \mathrm{or}\ 0 && \mathrm{otherwise}
         \end{aligned}
     \right.,
\end{equation}
where $k \sim U(0,1)$ is a random number and $[m]$ is the mask token.

\rev{
\textbf{Resampling} alters the sampling rate of a trajectory by resampling it at a different time interval. Given a trajectory $\mathcal{T}$ sampled with a time interval $t$, resampling can be performed to either increase (up-sample) or decrease (down-sample) the sampling rate:
}

\rev{
\textit{Down-sampling:} To decrease the sampling rate, a new trajectory $\mathcal T'$ can be obtained by sampling $\mathcal T$ with time interval $t'$, where $t'>t$, thus yielding a shorter trajectory. During down-sampling, trajectory points within each target interval $[t'_i, t'_{i+1}]$ are aggregated using several possible methods. The simplest approach is to select either the first or last point in each interval. More sophisticated methods include calculating the mean longitude and latitude of all points in the interval, taking the median position, or computing a time-weighted average where points are weighted by their temporal distance from the target timestamp.
}

\rev{
\textit{Up-sampling:} To increase the sampling rate, a new trajectory $\mathcal T'$ can be obtained by sampling $\mathcal T$ with time interval $t'$, where $t'<t$. New points between existing points can be generated using various interpolation techniques, such as linear interpolation.
}

\rev{
\textbf{Simplification} eliminates redundant points from a trajectory, preserving the overall shape of the trajectory by selecting key points that define its most significant features, or reduce the number of points in dense areas where multiple points carry redundant information.
The core idea of simplification is to identify segments of a trajectory that can be represented by fewer points while maintaining the essential information of the trajectory, in which geometric-based and density-based solutions~\cite{DBLP:journals/tvcg/VrotsouJNFSMAA15} can be used.
}

\textbf{Sequence of Input Trajectory Features} can be obtained by applying one or multiple preprocessing operations on one trajectory $\mathcal T$:
\begin{equation}
    \boldsymbol X_\mathcal T=\langle x_1, x_2, \dots, x_N \rangle,
\label{eq:sequence-of-input-trajectory-features}
\end{equation}
where $x_i$ is one step of processed features.

\subsection{Model}
\label{sec:model}
The model components constitute the majority of the learnable elements in a method for the pre-training of trajectory embeddings. The model components in UniTE include the feature embedder, encoder, decoder, and embedding postprocessor. Further, different instances of the components are provided; for example, different encoder and decoder component instances are provided for different neural architectures; see Figure~\ref{fig:pipeline}.
Among these, instances of the encoder component are responsible for mapping preprocessed trajectories to an embedding space. Optionally, a corresponding decoder component instance reconstructs trajectories from the embeddings.

\subsubsection{Feature Embedder}
Instances of the feature embedder component focus on mapping a preprocessed feature of a trajectory point into the embedding space. This allows for the establishment of information in trajectory points, facilitating the modeling of correlations between trajectory points in subsequent encoders and decoders.
Below we introduce the different instances of the feature embedder component.

\textbf{FC embedder} employs a fully-connected (FC) network to convert a feature $x$ into a $d$-dimensional embedding $\boldsymbol e$. The process for a single-layer FC embedder is described by:
\begin{equation}
    \mathrm{FC}(x) = g(\boldsymbol W x+\boldsymbol b),
\end{equation}
where $\boldsymbol W$ and $\boldsymbol b$ represent the embedder's weight and bias and $g$ is a non-linear activation function. To increase the embedding capacity, multiple layers of FC embedders can be stacked.

\textbf{Index-fetching embedder}, also known as \textit{lookup embedder}, forms a $d$-dimensional embedding vector for each unique token identified by the tokenization operation mentioned in Section~\ref{sec:feature-mapping}. This effectively yields an embedding matrix, with each row corresponding to the embedding of a specific token. The embedding vector for a discrete token $x$ is retrieved from the $x$-th row of the matrix.

\textbf{Word2vec embedder} pre-trains embedding vectors for a set of discrete tokens using word2vec~\cite{DBLP:conf/nips/MikolovSCCD13,Mikolov2013Efficient}. It requires sequences of discrete tokens as training data, where the sequences capture contextual correlations between the tokens. Such sequences can be derived directly from trajectories. The sequences of road segments obtained by map-matching trajectories is an example. Alternatively, they can be generated through algorithms like random walks in the road network. After word2vec training, this embedder functions similarly to the index-fetching embedder, with each discrete token being assigned a pre-trained embedding vector.

\textbf{Fourier embedder} utilizes learnable Fourier features~\cite{DBLP:conf/nips/TancikSMFRSRBN20} to transform a continuous feature $x$ into a $d$-dimensional embedding $\boldsymbol e$. The transformation is given as follows:
\begin{equation}
    \mathrm{Fourier}(x) = \frac{1}{\sqrt{d}}[\cos \boldsymbol wx \parallel \sin \boldsymbol wx],
\end{equation}
where $\boldsymbol w\in \mathbb R^{d/2}$ serves as a learnable mapping vector, $\parallel$ denotes vector concatenation. This embedding technique uses the periodic nature of trigonometric functions to preserve periodicity in features.

\textbf{Sequence of Embedding Vectors} is generated by applying one or more instances of embedder component on the sequence $\boldsymbol X_\mathcal T$ of input trajectory features calculated in Equation~\ref{eq:sequence-of-input-trajectory-features}. This resulting sequence of embedding vectors is as follows:
\begin{equation}
    \boldsymbol E_{\mathcal T} = \langle \boldsymbol e_1, \boldsymbol e_2, \dots, \boldsymbol e_N \rangle,
\end{equation}
where each $\boldsymbol e_i$ represents an embedding vector created by instances of embedder component. Subsequently, the sequence $\boldsymbol E_{\mathcal T}$ can be processed by instances of encoder and decoder components.

\subsubsection{RNN-based Encoder and Decoder}
\label{sec:rnn}
RNNs~\cite{DBLP:journals/cogsci/Elman90} are particularly adept at handling sequential data, making them well-suited for analyzing trajectories. In an RNN-based encoder, a sequence $\boldsymbol E_\mathcal T$ of embedding vectors is processed, and its final hidden state is considered as the embedding $\boldsymbol z_{\mathcal T}$ of the corresponding trajectory $\mathcal T$. This process can be expressed as follows:
\begin{equation}
    \boldsymbol z_{\mathcal T} = \mathrm{RNN}(\boldsymbol E_{\mathcal T})
\end{equation}

Similarly, in an RNN-based decoder, the trajectory embedding $\boldsymbol z_{\mathcal T}$ is used to reconstruct the trajectory sequence, typically in an auto-regressive manner. This process is defined as follows:
\begin{equation}
    \widehat{\mathcal T} = \mathrm{RNN}(\boldsymbol z_{\mathcal T}),
\end{equation}
where $\boldsymbol z_{\mathcal T}$ acts as the initial hidden state for the RNN decoder.
Both the encoder and decoder networks can employ one of three RNN variants: vanilla RNN, LSTM, or GRU, which differ in how they process each input step of trajectory features.

\textbf{Vanilla RNN} comprises an input, a hidden, and an output layer that can be extended in the temporal dimension. The functioning of a vanilla RNN at time $t$ is given as follows:
\begin{equation}
\begin{aligned} 
   h_t&=g_1(\boldsymbol W \boldsymbol e_t+\boldsymbol U \boldsymbol h_{t-1}+\boldsymbol b)\\
   y_t&= g_2(\boldsymbol V \boldsymbol h_t),
\end{aligned}
\end{equation}
where $\boldsymbol e_t$ is the input embedding vector of the input layer, $\boldsymbol h_t$ is the output of the hidden layer, $\boldsymbol y_t$ is the output of the output layer, $g_1$ and $g_2$ are non-linear activation functions, and $\boldsymbol W$, $\boldsymbol U$, $\boldsymbol V$, and $\boldsymbol b$ are weights and biases.

An RNN variant, the \textbf{LSTM}~\cite{hochreiter1997long} neural architecture addresses the issues of exploding or vanishing gradients encountered in vanilla RNNs. LSTM introduces three gates to regulate the flow of information: a forget gate $f$, an input gate $i$, and an output gate $o$. The formulations of these gates at step $t$ are as follows:
\begin{equation}
\begin{aligned} 
    f_{t}&=\sigma\left(\boldsymbol W_{f} \boldsymbol e_{t}+\boldsymbol U_{f} \boldsymbol h_{t-1}+\boldsymbol b_{f}\right) \\
    i_{t}&=\sigma\left(\boldsymbol W_{i} \boldsymbol e_{t}+\boldsymbol U_{i} \boldsymbol h_{t-1}+\boldsymbol b_{i}\right) \\
     o_{t}&=\sigma\left(\boldsymbol W_{o} \boldsymbol e_{t}+\boldsymbol U_{o} \boldsymbol h_{t-1}+\boldsymbol b_{o}\right),
\end{aligned}
\end{equation}
where $\sigma$ denotes the Sigmoid activation function. Additionally, a new cell state $\boldsymbol c_t$ is introduced to retain historical information up to the current time step and nonlinearly pass information to the hidden state $\boldsymbol h_t$. The cell state $\boldsymbol c_t$ and hidden state $\boldsymbol h_t$ are computed using the following equations:
\begin{equation}
\begin{aligned} 
\boldsymbol c_{t}&=f_{t} \odot \boldsymbol c_{t-1}+i_{t} \odot \tilde{\boldsymbol c}_{t} \\
    \boldsymbol h_{t}&=o_{t} \odot \tanh\left(\boldsymbol c_{t}\right),
\end{aligned}\label{eq:three_gates}
\end{equation}
where $\odot$ is the element-wise product, $\tanh$ is the Tanh activation function, $\boldsymbol c_{t-1}$ is the cell state at the previous time step, and $\tilde{\boldsymbol c}_{t}$ is the candidate memory cell computed as follows:
\begin{equation}
\tilde{\boldsymbol c}_{t}=\tanh\left(\boldsymbol W_{c} \boldsymbol e_{t}+\boldsymbol U_{c} \boldsymbol h_{t-1}+\boldsymbol b_{c}\right)
\end{equation}

The \textbf{GRU}~\cite{DBLP:journals/corr/ChungGCB14} neural architecture is a simpler RNN variant than the LSTM. It incorporates an update gate $z$ and a reset gate $r$. Unlike the LSTM, GRU combines the cell state and the output into a single state $\boldsymbol h$ without introducing additional memory cells. The update gate $z_t$ determines the amount of information that the current state $\boldsymbol h_t$ is to retain from the previous state $\boldsymbol h_{t-1}$ and how much new information it is to receive from the candidate state $\tilde{\boldsymbol h}_{t}$. The reset gate $r_t$ decides whether the calculation of $\tilde{\boldsymbol h}_{t}$ depends on $\boldsymbol h_{t-1}$. The operation of the GRU at step $t$ is expressed as follows:
\begin{equation}
\begin{aligned} 
    z_{t}&=\sigma\left(\boldsymbol W_{z} \boldsymbol e_{t}+\boldsymbol U_{z} \boldsymbol h_{t-1}+\boldsymbol b_{z}\right) \\
    r_{t}&=\sigma\left(\boldsymbol W_{r} \boldsymbol e_{t}+\boldsymbol U_{r} \boldsymbol h_{t-1}+\boldsymbol b_{r}\right) \\   \tilde{\boldsymbol h}_{t}&=\tanh\left(\boldsymbol W_{h} \boldsymbol e_{t}+\boldsymbol U_{h} (r_t\odot \boldsymbol h_{t-1})+\boldsymbol b_{h}\right) \\
    \boldsymbol h_{t}&=(1-z_{t})\odot \boldsymbol h_{t-1} + z_{t}\odot\tilde{\boldsymbol h}_{t}
\end{aligned}
\end{equation}

In an RNN-based encoder, the final hidden state $\boldsymbol h_N$ is considered the encoded embedding vector $\boldsymbol z_\mathcal T$ of the input sequence $\boldsymbol E_\mathcal T$, with $N$ representing the sequence length. Conversely, in an RNN-based decoder, the sequence of outputs $\langle y_1, y_2, \dots, y_N \rangle$ is viewed as the reconstructed trajectory $\widehat{\mathcal T}$.

\subsubsection{Transformer-based Encoder and Decoder}
The advanced self-attention mechanism in Transformers~\cite{vaswani2017attention} enables them to understand complex spatiotemporal relationships in trajectories. A Transformer-based encoder processes a sequence $\boldsymbol{E}_\mathcal{T}$ of embedding vectors and generates a memory sequence $\boldsymbol{M}_\mathcal{T}$ of the same length. To derive the embedding $\boldsymbol{z}_{\mathcal{T}}$, a pooling operation is applied to $\boldsymbol{M}_\mathcal{T}$. This process is formulated as follows:
\begin{equation}
\begin{split}
    \boldsymbol{M}_{\mathcal{T}} &= \mathrm{Transformer}(\boldsymbol{E}_\mathcal{T})\\
    \boldsymbol{z}_{\mathcal{T}} &= \mathrm{Pool}(\boldsymbol{M}_{\mathcal{T}})
\end{split}
\label{eq:transformer-encoder}
\end{equation}

Next, a Transformer-based decoder aims to reconstruct the trajectory sequence from $\boldsymbol{M}_\mathcal{T}$ by using $\boldsymbol{M}_\mathcal{T}$ as the query in the Attention mechanism. Thus, we get:
\begin{equation}
    \widehat{\mathcal{T}} = \mathrm{Transformer}(\boldsymbol{M}_\mathcal{T}, \mathcal{T}_\text{src}),
    \label{eq:transformer-decoder}
\end{equation}
where $\mathcal{T}_\text{src}$ is the source trajectory guiding the generation of $\widehat{\mathcal{T}}$.

Multi-head attention is the key component of the conventional transformer architecture, enabling the network to focus on every token within the input sequence. To implement multi-head attention, the input is first projected to three matrices, query $\boldsymbol{Q}$, key $\boldsymbol{K}$, and value $\boldsymbol{V}$, through linear transformations. This process in Equations~\ref{eq:transformer-encoder} and \ref{eq:transformer-decoder} is formulated as follows:
\begin{equation}
\begin{split}
    [\boldsymbol{Q}_i, \boldsymbol{K}_i, \boldsymbol{V}_i] &= \boldsymbol{E}_\mathcal{T}[\boldsymbol{W}_i^Q, \boldsymbol{W}_i^K, \boldsymbol{W}_i^V]\\
    [\boldsymbol{Q}_i, \boldsymbol{K}_i, \boldsymbol{V}_i] &= [\boldsymbol{M}_\mathcal{T} \boldsymbol{W}_i^Q, \boldsymbol{E}_{\mathcal{T}_\text{src}} \boldsymbol{W}_i^K, \boldsymbol{E}_{\mathcal{T}_\text{src}} \boldsymbol{W}_i^V],
\end{split}
\end{equation}
where $\boldsymbol{E}_{\mathcal{T}_\text{src}}$ is the embedding sequence of $\mathcal{T}_\text{src}$, $\boldsymbol{W}_i^Q, \boldsymbol{W}_i^K \in \mathbb{R}^{d \times d_{QK}}$, and $\boldsymbol{W}_i^V \in \mathbb{R}^{d \times d_V}$ are the transformation matrices. Queries $\boldsymbol{Q}_i$ and keys $\boldsymbol{K}_i$ have the same dimensionality $d_{QK}$, while values $\boldsymbol{V}_i$ have dimensionality $d_V$. In practice, we usually set $d_{QK} = d_V = d$. 

The output matrix of the attention mechanism of the $i$-th head is then obtained as follows:
\begin{equation}
    \mathrm{Attention}(\boldsymbol{Q}_i, \boldsymbol{K}_i, \boldsymbol{V}_i) = \mathrm{softmax}\left(\frac{\boldsymbol{Q}_i \boldsymbol{K}_i^\mathrm{T}}{\sqrt{d}}\right)\boldsymbol{V}_i,
\end{equation}
where $\mathrm{softmax}$ is row-wise softmax normalization. To incorporate multiple aspects of correlation, multi-head attention first concatenates the output of multiple attention heads into a long vector, and then multiplies a weight matrix $\boldsymbol{W}^O$ with a fully-connected network to obtain the final output. This process is formulated as follows:
\begin{equation}
\begin{split} 
    \mathrm{MultiHeadAtt}(X) = \mathrm{Concat}(
    &\mathrm{Attention}(\boldsymbol{Q}_1, \boldsymbol{K}_1, \boldsymbol{V}_1),
    \ldots,\\
    &\mathrm{Attention}(\boldsymbol{Q}_h, \boldsymbol{K}_h, \boldsymbol{V}_h))\boldsymbol{W}^O,
\end{split}
\end{equation}
where $h$ is the number of heads. 

Following the multi-head attention module, the Transformer extracts deeper features using a Feed Forward network that usually includes two linear layers and a non-linear activation function, mapping data to high-dimensional spaces and then to low-dimensional spaces. This process is formulated as follows:
\begin{equation}
    \mathrm{FFN}(x) = \boldsymbol{W}_2(g(\boldsymbol{W}_1 x + \boldsymbol{b}_1)) + \boldsymbol{b}_2,
\end{equation}
where $\boldsymbol{W}_1, \boldsymbol{W}_2$, $\boldsymbol{b}_1$, and $\boldsymbol{b}_2$ represent the weights and biases of the linear layers, and $g$ is a non-linear activation function.

Moreover, Transformers use residual connections in each module separately. That is, the output of each transformer layer is:
\begin{equation}
\begin{split} 
    \boldsymbol{z}_i^{multi} &= \mathrm{LayerNorm}(\boldsymbol{z}_{i-1} + \mathrm{MultiHeadAtt}(\boldsymbol{z}_{i-1})) \\
    \boldsymbol{z}_i &= \mathrm{LayerNorm}(\boldsymbol{z}_i^{multi} + \mathrm{FFN}(\boldsymbol{z}_i^{multi})),
\end{split}
\end{equation}
where $\boldsymbol{z}_{i-1}$ and $\boldsymbol{z}_i$ represent the input and output of the $i$-th transformer layer, respectively.

In a Transformer-based encoder, the output of the last transformer layer is regarded as the memory sequence $\boldsymbol{M}_\mathcal{T}$. Conversely, in a Transformer-based decoder, the output of the last transformer layer is fed into a fully-connected prediction module to produce the reconstructed trajectory $\widehat{\mathcal{T}}$.

\subsubsection{CNN-based Encoder and Decoder}
Convolutional Neural Networks (CNNs)~\cite{DBLP:journals/pr/FukushimaM82,DBLP:conf/cvpr/CiresanMS12} are well-suited for capturing intricate spatial patterns, making them suitable for analyzing trajectories with complex spatial features. In the context of trajectory analysis, a CNN-based encoder processes the image representation $\boldsymbol{T}$ of a trajectory $\mathcal{T}$, which has been preprocessed by the pixelation operation as discussed in Section~\ref{sec:coordinate-mapping}. By employing multi-layer CNNs in conjunction with fully-connected and pooling layers, the encoder produces an embedding $\boldsymbol{z}_{\mathcal{T}}$ as follows:
\begin{equation}
    \boldsymbol{z}_{\mathcal{T}} = \mathrm{CNN}(\boldsymbol{T})
    \label{eq:cnn-encoder}
\end{equation}

Next, the CNN-based decoder reconstructs the trajectory image $\boldsymbol{T}$ from its embedding $\boldsymbol{z}_{\mathcal{T}}$ by essentially reversing the encoding process. This decoding operation mirrors the structure of the encoder:
\begin{equation}
    \widehat{\boldsymbol{T}} = \mathrm{CNN}(\boldsymbol{z}_{\mathcal{T}})
\end{equation}

To elaborate further, the CNN network in Equation~\ref{eq:cnn-encoder} comprising two convolution layers, a pooling layer, and a fully-connected layer can be defined as follows:
\begin{equation}
    \mathrm{CNN}(\boldsymbol{T}) = \mathrm{FC}(\mathrm{Pool}(\boldsymbol W_2 * g(\boldsymbol W_1 * \boldsymbol{T}))),
\end{equation}
where $*$ denotes the convolution operation, $\mathrm{Pool}$ is the pooling operation, $\mathrm{FC}$ represents the fully-connected layer, $\boldsymbol W_1$ and $\boldsymbol W_2$ are convolution kernels, and $g$ is a non-linear activation function.

\subsubsection{ODE-based Encoder and Decoder}
The NeuralODE family~\cite{DBLP:conf/nips/ChenRBD18} represents a novel approach to capturing the continuous dynamics of data, thus offering a new perspective on trajectory modeling. Building upon the foundations laid by the RNN-based encoder discussed in Section~\ref{sec:rnn}, the ODE-based encoder updates its hidden states not only at each step of an input embedding $\boldsymbol e_i$, but also between these steps through the use of an ODE solver. The update process is described as follows:
\begin{equation}
    \boldsymbol h_{i-1}' = \mathrm{ODESolve}(\boldsymbol h_{i-1}, (t_{i-1},t_i)),
\end{equation}
where $\boldsymbol h_{i-1}'$ is the newly updated hidden state, which is then further processed by the RNN cell. Similar to the RNN-based encoder and decoder, the ODE-based encoder compresses the embedding sequence $\boldsymbol E_{\mathcal T}$ into a trajectory embedding $\boldsymbol z_{\mathcal T}$, and the ODE-based decoder reconstructs trajectory $\widehat{\mathcal T}$ from $\boldsymbol z_{\mathcal T}$.

\textbf{CDE-based Encoder} is an expanded variant of the ODE-based encoder based on NeuralCDE~\cite{DBLP:conf/nips/KidgerMFL20}. It employs a spline $\boldsymbol T$, derived from the spline operation introduced in Section~\ref{sec:augmentation}, and performs integration over this spline across time from $t_1$ to $t_N$. This integration, carried out with a specially parameterized CDE integral kernel, produces the trajectory embedding $\boldsymbol z_{\mathcal T}$.

\subsubsection{Embedding Postprocessor}
The embeddings calculated by the instances of encoder component can be optionally transformed by utilizing the embedding postprocessor component. Below we present different instances of the component.

\textbf{Variational postprocessor} maps the output of an encoder to a multi-dimensional Gaussian space. Given an embedding $\boldsymbol z$, it uses two fully-connected networks to compute the mean and variance of a Gaussian distribution:
\begin{equation}
    \boldsymbol \mu = \boldsymbol W_{\mu}\boldsymbol z + \boldsymbol b_{\mu},
    \boldsymbol \sigma = \boldsymbol W_{\sigma}\boldsymbol z + \boldsymbol b_{\sigma},
\end{equation}
where $\boldsymbol W_{\mu}$, $\boldsymbol W_{\sigma}$, $\boldsymbol b_{\mu}$, and $\boldsymbol b_{\sigma}$ are weights and biases. The processed embedding is then obtained from the computed Gaussian distribution:
\begin{equation}
    \boldsymbol z' \sim \mathcal N(\boldsymbol z'|\boldsymbol \mu, \boldsymbol \sigma)
    \label{eq:sampled-embedding}
\end{equation}

\textbf{NF postprocessor}, also called the normalizing flows postprocessor, transforms simple probability distributions into more complex ones through a series of invertible and differentiable mappings. This method leverages the concept of normalizing flows to enhance the expressiveness of probabilistic models.
At its core, a normalizing flow (NF)~\cite{DBLP:journals/jmlr/PapamakariosNRM21} consists of a sequence of invertible functions $g_1, g_2, \dots, g_K $. Each of these functions, also called transformations, is designed to be bijective and differentiable, ensuring that the overall transformation remains invertible and that the change in probability density can be tracked through the Jacobian determinant.

Given an initial simple distribution, typically a multivariate Gaussian distribution $\mathcal N(\boldsymbol z' | \boldsymbol \mu, \boldsymbol \sigma)$ from the above variational postprocessor, the NF Postprocessor applies these transformations sequentially to derive a more complex distribution. This process can be stated as follows:
\begin{equation}
\boldsymbol z'' = g_K \circ g_{K-1} \circ \cdots \circ g_1(\boldsymbol z'),
\end{equation}
where $\boldsymbol z'$ is a sample drawn from the initial Gaussian distribution. The transformed variable $\boldsymbol z''$ now follows a more complex distribution that is potentially better suited for subsequent downstream tasks.

\subsection{Pre-training Process}
\label{sec:pre-training-process}
The pre-training process is a vital component of the UniTE pipeline, ensuring that the learnable parameters of models are trained in a self-supervised manner, yielding trajectory embeddings that are useful for downstream tasks. We consider two key components that support the pre-training process: the loss function and pre-trainer components.

\subsubsection{\rev{Loss Function}}
\label{sec:loss-function}
\rev{
Loss function component instances are designed to optimize the learnable parameters by penalizing deviations from desired behaviors. UniTE includes three types of loss functions: reconstruction, contrastive, and regularization loss functions.
}

\textbf{Reconstruction Loss} quantifies the difference between the original input trajectories and their reconstructed counterparts as produced by the decoder. This metric is crucial for ensuring that both the encoder and decoder are effectively capturing and retaining important information contained in trajectories.

The particular function used for calculating reconstruction loss varies based on the characteristics of the features being reconstructed.

\textit{MSE and MAE loss}, i.e., Mean Squared Error and Mean Absolute Error loss, are employed frequently for monitoring the reconstruction of continuous features. Given a reconstructed feature $\hat x$ and the ground truth $x$, these losses are defined as follows:
\begin{equation}
    \mathrm{MSE}(x, \hat x) = (x - \hat x)^2, 
    \mathrm{MAE}(x, \hat x) = |x - \hat x|
\end{equation}

\textit{Cross-entropy loss} is used for the reconstruction of discrete tokens. It involves comparing the predicted probability distribution of tokens $\hat p(x)$ against the actual token $x$ and is defined as follows:
\begin{equation}
    \mathrm{CE}(\hat p(x),x) = -\hat p(x)_x + \log(\sum_c^C \exp(\hat p(x)_c)),
\end{equation}
where $C$ is the total count of unique tokens.

\textit{Distance loss} is tailored for the reconstruction of spatial data. It computes the loss based on the geometric distance between predicted coordinates $\hat{l_i}$ and the ground truth $l_i$, often using the shortest path on the Earth's surface or on the road network as the distance function.

\textbf{Contrastive Loss} is used to refine embeddings by differentiating between similar (positive) and dissimilar (negative) pairs of data points. This technique aims to group embeddings of similar trajectories closely together and separate dissimilar trajectories.

\textit{InfoNCE loss}~\cite{oord2018representation} enhances a model by maximizing the mutual information between positive pairs in contrast with a selection of negative samples. Given a set of trajectories $\{\mathcal T_1,\mathcal T_2, \dots, \mathcal T_B\}$ and a target trajectory $\mathcal T_i$, a positive pair of embeddings $\boldsymbol z_{\mathcal T_i}$ and $\boldsymbol z'_{\mathcal T_i}$ is calculated, usually representing two different augmentations of $\mathcal T$. Meanwhile, the embeddings $\boldsymbol z'_{\mathcal T_j}, j\neq i$ for the other trajectories are considered as negative samples. The goal of the InfoNCE loss is to effectively distinguish positive and negative samples, formulated as follows:
\begin{equation}
    \mathcal L_\mathrm{InfoNCE} = -\log \frac{\exp(\boldsymbol z_{\mathcal T_i}{\boldsymbol z'_{\mathcal T_i}}^\top/\tau)}{\sum_{j=1}^B \exp(\boldsymbol z_{\mathcal T_i}{\boldsymbol z'_{\mathcal T_j}}^\top/\tau)},
\end{equation}
where $\tau$ is the temperature parameter.

\textit{MEC loss}~\cite{liu2022self} utilizes the principle of maximum entropy from information theory to direct the learning of general trajectory embeddings. Given a set of trajectories $\{\mathcal T_1,\mathcal T_2, \dots, \mathcal T_B\}$, two sets of their embeddings $\boldsymbol Z^{(1)}\in \mathbb R^{B\times d}$ and $\boldsymbol Z^{(2)}\in \mathbb R^{B\times d}$ are produced through different preprocessing, embedding, and encoding steps. The MEC loss is then defined as follows:
\begin{equation}
    \mathcal L_\mathrm{MEC} = \frac{B+d}{2} \log \det (\boldsymbol I_B + \frac{d}{B\epsilon^2} \boldsymbol Z^{(1)} {\boldsymbol Z^{(2)}}^\top),
\end{equation}
where $\boldsymbol I_B$ is an identity matrix of dimensionality $B$ and $\epsilon$ is the upper bound of the decoding error.

\textbf{Regularization Loss} is employed to prevent overfitting by encouraging the model to learn more generalized embeddings. These losses are typically applied to the learned embeddings or the latent states of models, serving to constrain the complexity of the model and improve its generalization capabilities.

\textit{L1 and L2 loss} are widely used for regularization purposes, and are also known as Lasso and Ridge regularization, respectively. They impose penalties according to the magnitude of the parameters. Given a trajectory embedding $\boldsymbol z_{\mathcal T}$, these losses are calculated as follows:
\begin{equation}
    \mathcal L_\mathrm{L1} = \sum_i |{\boldsymbol z_{\mathcal T}}_i|,~
    \mathcal L_\mathrm{L2} = \sqrt{\sum_i {{\boldsymbol z_{\mathcal T}}_i}^2},
\end{equation}
where ${\boldsymbol z_{\mathcal T}}_i$ represents the $i$-th dimension of $\boldsymbol z_{\mathcal T}$. The L1 loss encourages sparsity by driving many of the parameters to zero, while the L2 loss prevents large weights by penalizing the square of the magnitude of the parameters.

\textit{ELBO loss} is an essential component in the Variational Autoencoder (VAE) framework for regularizing the learned distribution of embeddings. The Evidence Lower Bound (ELBO) is used to approximate the likelihood of the data under the model. The ELBO loss consists of two main terms: a reconstruction loss and the Kullback-Leibler (KL) divergence.

Given a trajectory $\mathcal T$ and its corresponding embedding vector $\boldsymbol z_\mathcal T$, the ELBO loss is defined as follows:
\begin{equation}
    \mathcal L_\text{ELBO} = \mathbb{E}_{q(\boldsymbol z_\mathcal T|\mathcal T)} \left[ \log p(\mathcal T|\boldsymbol z_\mathcal T) \right] - \mathrm{KL}\left( q(\boldsymbol z_\mathcal T|\mathcal T) \parallel p(\boldsymbol z_\mathcal T) \right),
\end{equation}
where $q(\boldsymbol z_\mathcal T|\boldsymbol x)$ is the variational posterior, $p(\mathcal T|\boldsymbol z_\mathcal T)$ is the likelihood, and $p(\boldsymbol z_\mathcal T)$ is the prior distribution of the embedding. The first term, $\mathbb{E}_{q(\boldsymbol z_\mathcal T|\mathcal T)} \left[ \log p(\mathcal T|\boldsymbol z_\mathcal T) \right]$, represents the reconstruction loss, which ensures that the model can accurately reconstruct the trajectory from the embedding vector. The second term, $\mathrm{KL}\left( q(\boldsymbol z_\mathcal T|\mathcal T) \parallel p(\boldsymbol z_\mathcal T) \right)$, is the KL divergence, which regularizes the latent space to match a prior distribution (commonly a standard normal distribution).

In summary, the L1 and L2 loss functions are applied to the model parameters to enforce sparsity and prevent large weights, respectively, while the ELBO loss function in VAEs regularizes the learned embeddings by balancing reconstruction accuracy and the regularity of the latent space distribution.

\subsubsection{\rev{Pre-trainer}}
\label{sec:pre-trainer}
\rev{
Pre-trainer component instances play a crucial role in integrating various elements to effectively pre-train trajectory embeddings. UniTE includes three types of pre-trainers: generative, contrastive, and hybrid pre-trainers.
}

\textbf{Generative Pre-trainer} aims to enhance embeddings by determining some segments of trajectory data based on other segments. This process entails either reconstructing a trajectory from a modified or condensed form, or predicting future segments based on past segments.
Given a trajectory $\mathcal T$, one or several of the preprocessors mentioned in Section~\ref{sec:preprocessor} are utilized to generate a modified, augmented, or feature-enhanced version of the trajectory. This version is then processed through an encoder and a decoder. The encoder condenses the trajectory into an embedding, while the decoder reconstructs segments or the entire trajectory. The pre-trainer supervises the learnable parameters, applying the reconstruction loss metrics detailed in Section~\ref{sec:loss-function} to the reconstructed trajectory. This pre-training procedure is repeated for each trajectory $\mathcal T$ in the trajectory dataset $\mathbb T$.
\rev{Algorithm~\ref{alg:generative-pretrainer} presents the detailed procedures of the generative pre-trainer.}

\begin{algorithm}[t]
    \caption{\rev{Generative Pre-trainer}}
    \label{alg:generative-pretrainer}
    \begin{algorithmic}[1]
    \Require \rev{Trajectory dataset $\mathbb{T}$, preprocessors $\mathcal{P}$, encoder $f_{\text{enc}}$, decoder $f_{\text{dec}}$}
    \Ensure \rev{Trained encoder and decoder parameters}
    \For{\rev{each epoch}}
        \For{\rev{each trajectory $\mathcal{T} \in \mathbb{T}$}}
            \State \rev{Generate modified trajectory $\mathcal{T}' \gets p(\mathcal{T})$ for each preprocessor $p \in \mathcal{P}$}
            \State \rev{Compute embedding $\boldsymbol{z}_{\mathcal{T}} \gets f_{\text{enc}}(\mathcal{T}')$}
            \State \rev{Reconstruct trajectory $\widehat{\mathcal{T}} \gets f_{\text{dec}}(\boldsymbol{z}_{\mathcal{T}})$}
            \State \rev{Update $f_{\text{enc}}$ and $f_{\text{dec}}$ using $\mathcal{L} = \text{ReconstructionLoss}(\mathcal{T}, \widehat{\mathcal{T}})$}
        \EndFor
    \EndFor
    \end{algorithmic}
\end{algorithm}

\textbf{Contrastive Pre-trainer} aims to refine embeddings by differentiating between similar (positive) and dissimilar (negative) pairs of trajectories. This method is dependent on being able to generate meaningful positive and negative examples.
For a given trajectory $\mathcal T$, instances of preprocessor component as those covered in Section~\ref{sec:preprocessor} are employed to produce multiple trajectory augmentations. These augmented versions are then fed into the encoders to extract their embeddings. Following this, the pre-trainer applies the contrastive loss metrics specified in Section~\ref{sec:loss-function} to these embeddings to guide the learning of parameters. This process is performed iteratively for every trajectory $\mathcal T$ in $\mathbb T$. 
\rev{Algorithm~\ref{alg:contrastive-pretrainer} presents the detailed procedures of the contrastive pre-trainer.}

\begin{algorithm}[t]
    \caption{\rev{Contrastive Pre-trainer}}
    \label{alg:contrastive-pretrainer}
    \begin{algorithmic}[1]
    \Require \rev{Trajectory dataset $\mathbb{T}$, preprocessors $\mathcal{P}$, encoder $f_{\text{enc}}$, number of augmentations $K$}
    \Ensure \rev{Trained encoder parameters}
    \For{\rev{each epoch}}
        \For{\rev{each trajectory $\mathcal{T} \in \mathbb{T}$}}
            \State \rev{Generate $K$ augmented versions $\{\mathcal{T}'_1, ..., \mathcal{T}'_K\}$ using preprocessors $\mathcal{P}$}
            \State \rev{Compute positive embeddings $\boldsymbol{Z}^+ = \{f_{\text{enc}}(\mathcal{T}'_i)\}_{i=1}^K$}
            \State \rev{Sample negative trajectories $\{\mathcal{T}_j\}_{j\neq i}$ from $\mathbb{T}$}
            \State \rev{Compute negative embeddings $\boldsymbol{Z}^- = \{f_{\text{enc}}(\mathcal{T}_j)\}_{j\neq i}$}
            \State \rev{Update $f_{\text{enc}}$ using $\mathcal{L} = \text{ContrastiveLoss}(\boldsymbol{Z}^+, \boldsymbol{Z}^-)$}
        \EndFor
    \EndFor
    \end{algorithmic}
\end{algorithm}

\textbf{Hybrid Pre-trainer} amalgamates the methodologies of generative and contrastive pre-trainers, aiming to benefit from both reconstruction and discrimination tasks. It adheres to the procedures established by the generative and contrastive pre-trainers, while combining these through a loss that is a weighted summation of the losses from each pre-trainer type.

\subsection{Downstream Adapter}
\label{sec:downstream-adaptor}
The downstream adapter functions as an intermediary in-between pre-trained trajectory embeddings and their application in downstream tasks. It customizes the universal embeddings resulting from the pre-training to suit particular tasks, potentially enhancing the effectiveness of embeddings through fine-tuning. Moreover, the adapter offers a standardized and detailed approach for assessing and benchmarking the performance of different pre-training methods across different downstream tasks.

\subsubsection{Destination Prediction Adapter}
This adapter is dedicated to the task of forecasting the destination of a trajectory. When calculating a trajectory $\mathcal T$'s embedding $\boldsymbol z_{\mathcal T}$, the last $L$ points of $\mathcal T$ are omitted. A fully-connected network then uses this embedding to predict the destination point's road segment $s_N$. The cross-entropy loss is applied to the predicted segment to refine the learnable parameters.

\textbf{Evaluation metrics} for this adapter include Acc@1, Acc@5, Recall, and F1. Acc@1 and Acc@5 measure the percentages of correct top-1 and top-5 predictions, respectively. The Recall and F1 metrics calculate the recall and F1 scores for each class label---in this case, each road segment---and average the scores across all classes.

\subsubsection{Arrival Time Estimation Adapter}
Designed for estimating the arrival time at a destination from a trajectory, this adapter also omits the final $L$ points of a trajectory $\mathcal T$ when calculating the embedding $\boldsymbol z_{\mathcal T}$. A fully-connected network employs the embedding to forecast the destination point's arrival time $t_N$. The Mean Absolute Error (MAE) loss or the Mean Squared Error (MSE) loss is used on the prediction to fine-tune parameters.

\textbf{Evaluation metrics} for this adapter include MAE, Root Mean Squared Error (RMSE), and Mean Absolute Percentage Error (MAPE).

\subsubsection{Trajectory Classification Adapter}
This adapter is designed to predict the class label of a trajectory, such as its driver ID. Given a trajectory $\mathcal{T}$, its entire sequence is used to compute its embedding vector $\boldsymbol{z}_\mathcal{T}$. A fully-connected network then uses $\boldsymbol{z}_\mathcal{T}$ to predict the class label of $\mathcal{T}$. The model's parameters are fine-tuned using the cross-entropy loss.

\textbf{Evaluation Metrics} for this adapter are the same as for the destination prediction adapter, as both perform classification.

\subsubsection{Similar Trajectory Search Adapter}
This adapter targets the unsupervised task of identifying the trajectory in a set of trajectories that is most similar to a target trajectory. Given a target trajectory $\mathcal T_t$ and a set of candidate trajectories $\{\mathcal T_1, \mathcal T_2, \dots, \mathcal T_B\}$, the similarity between the target's embedding $\boldsymbol z_{\mathcal T_t}$ and each candidate's embedding $\boldsymbol z_{\mathcal T_i}$ is calculated using cosine similarity. The trajectory with the highest similarity is deemed the most similar trajectory.

\textbf{Evaluation Metrics} for this adapter include the same set of metrics as for the destination prediction adapter, as this adapter can be regarded as a classification task.

\subsubsection{Training Strategy}
\label{sec:training-strategy}
There are two sources of supervision in the pipeline: the pre-training process covered in Section~\ref{sec:pre-training-process} and the fine-tuning process guided by the adapters just presented. We incorporate three training strategies into the pipeline:
\rev{
(1) \textbf{wo finetune} (without finetune), which reduces the emphasis on fine-tuning embedding models. After their pre-training, the parameters in the embedding models are kept fixed;
(2) \textbf{wo pretrain} (without pre-train), which bypasses the pre-training of embedding models. The parameters in the embedding models are initialized randomly and updated directly using the task-specific loss function in the adapters;
and (3) \textbf{full}, which includes both pre-training and fine-tuning. The parameters in the embedding models are first learned through their pre-training processes and then further refined using the task-specific loss functions in the adapters.
}

It is important to note that the parameters in the prediction networks are always updated using the task-specific loss function in the adapters, regardless of the strategy.

\begin{table*}
\centering
\caption{Implementation of existing methods for the pre-training of trajectory embeddings in UniTE.}
\resizebox{1.0\linewidth}{!}{
\begin{tabular}{c|cccccc}
\toprule
\multirow{2}{*}{Methods} & \multirow{2}{*}{Preprocessor} & Feature & Encoder and & Embedding & Loss & \multirow{2}{*}{Pre-trainer} \\
& & Embedder & Decoder & Postprocessor & Function & \\
\midrule
FVTI & Sliding-Window+Tokenization & Word2vec & - & - & - & - \\
GCM & Tokenization & Word2vec & - & - & - & - \\
POI2Vec & Tokenization & Word2vec & - & - & - & - \\
TALE & Tokenization & Word2vec & - & - & - & - \\
\midrule
CTLE & Augmentation & Index-fetching+FC & Transformer & - & Reconstruction & Generative \\
Toast & Augmentation & Index-fetching & Transformer & - & Reconstruction+Contrastive & Hybrid \\
\midrule
DTC & Tokenization & Index-fetching & RNN & - & Reconstruction & Generative \\
trajectory2vec & Sliding-Window & FC & RNN & - & Reconstruction & Generative \\
TremBR & Map-matching & Word2vec+FC & RNN & - & Reconstruction & Generative \\
CAETSC & Pixelation & FC & CNN & - & Reconstruction & Generative \\
\midrule
GM-VSAE & Tokenization & Index-fetching & RNN & Variational & Reconstruction+Regularization & Generative \\
TrajODE & Normalization & FC & RNN+ODE & Variational+NF & Reconstruction+Regularization & Generative \\
\midrule
t2vec & Augmentation+Tokenization & Index-fetching & RNN & - & Reconstruction & Generative \\
Robust DAA & Pixelation & FC & CNN & - & Reconstruction & Generative \\
TrajectorySim & Augmentation+Tokenization & Index-fetching & RNN & - & Reconstruction & Generative \\
\midrule
PreCLN & Map-matching+Augmentation & Index-fetching+FC & Transformer & - & Contrastive & Contrastive \\
TrajCL & Augmentation+Tokenization & Index-fetching+FC & Transformer & - & Contrastive & Contrastive \\
START & Map-matching+Augmentation & Index-fetching+FC & Transformer & - & Contrastive+Reconstruction & Hybrid \\
LightPath & Map-matching+Augmentation & Index-fetching & Transformer & - & Contrastive+Reconstruction & Hybrid \\
MMTEC & Map-matching+Spline & Index-fetching+FC & Transformer+CDE & - & Contrastive & Contrastive \\
\bottomrule
\end{tabular}
}

\label{tab:methods}
\end{table*}

\begin{table*}
\centering
\caption{\rev{Performance comparison of different approaches on the destination prediction task.}}
\begin{threeparttable}
\resizebox{1.0\linewidth}{!}{
    \begin{tabular}{c|cccc|cccc}
    \toprule
    Strategy & \multicolumn{8}{c}{wo finetune / wo pretrain / full} \\
    \midrule
    \multirow{2}{*}{Method} &\multicolumn{4}{c|}{Chengdu} & \multicolumn{4}{c}{Porto} \\ 
    \cline{2-9}
    & Acc@1(\%) & Acc@5(\%) & Recall(\%) & f1(\%) & 
    Acc@1(\%) & Acc@5(\%) & Recall(\%) & f1(\%) \\
    \midrule
    \rev{FVTI} & \rev{5.89/13.20/13.35} & \rev{15.64/27.04/27.83} & \rev{0.83/2.80/2.96} & \rev{0.51/2.16/2.23} & \rev{13.32/24.56/25.03} & \rev{27.86/49.69/50.98} & \rev{1.50/4.25/4.47} & \rev{1.05/3.28/3.31} \\
    GCM & 53.73/54.64/54.88 & 86.56/87.59/87.59 & 22.45/22.79/23.03 & 20.85/20.96/21.21 & 33.43/34.29/34.43 & 66.74/68.05/68.51 & 8.16/8.61/8.95 & 7.36/7.61/8.05 \\
    \rev{POI2Vec} & \rev{30.22/25.97/47.21} & \rev{52.49/48.66/76.13} & \rev{12.14/2.28/20.62} & \rev{11.77/2.01/18.12} & \rev{17.83/18.31/19.19} & \rev{36.67/38.13/40.12} & \rev{3.23/3.71/3.78} & \rev{2.33/2.85/2.91} \\
    \rev{TALE} & \rev{30.35/26.10/47.88} & \rev{52.81/49.02/77.26} & \rev{12.22/2.34/21.13} & \rev{11.85/2.08/18.45} & \rev{18.05/18.56/19.45} & \rev{37.01/38.45/40.89} & \rev{3.30/3.82/3.91} & \rev{2.42/2.92/3.02} \\
    \rev{CTLE} & \rev{18.93/26.43/35.20} & \rev{44.04/55.75/66.84} & \rev{2.54/2.62/4.58} & \rev{2.37/2.04/3.25} & \rev{10.56/18.56/22.56} & \rev{25.33/43.68/51.61} & \rev{0.83/2.44/2.75} & \rev{0.37/1.87/1.77} \\
    \rev{Toast} & \rev{20.81/27.42/37.02} & \rev{46.75/56.13/68.41} & \rev{2.89/2.50/5.80} & \rev{2.46/1.44/4.38} & \rev{19.21/20.17/27.68} & \rev{47.66/44.15/62.25} & \rev{3.14/2.31/4.74} & \rev{1.41/2.14/3.13} \\
    DTC & 6.56/55.43/54.69 & 16.86/87.87/86.44 & 0.13/23.56/23.31 & 0.03/21.40/21.27 & 4.02/38.26/36.02 & 10.29/73.72/70.86 & 0.12/10.53/8.90 & 0.01/8.84/7.87 \\
    trajectory2vec & 26.57/29.35/36.25 & 53.52/56.97/64.99 & 3.99/4.68/7.55 & 3.66/4.15/6.99 & 8.25/4.92/6.07 & 21.17/13.50/15.96 & 0.46/0.14/0.39 & 0.22/0.03/0.16 \\
    TremBR & 22.44/57.86/56.09 & 48.91/90.27/88.40 & 3.40/25.64/23.88 & 2.56/23.87/22.06 & 4.89/38.50/38.14 & 13.68/75.31/74.16 & 0.14/11.03/10.14 & 0.03/9.50/8.60 \\
    CAETSC & 18.55/24.38/16.71 & 50.35/60.50/43.17 & 4.73/7.07/2.59 & 3.23/5.22/1.43 & 15.99/34.20/28.74 & 36.19/67.22/58.29 & 1.75/8.91/4.71 & 1.04/7.70/3.46 \\
    GM-VSAE & 31.85/38.96/39.25 & 70.07/80.82/80.67 & 15.04/18.67/18.91 & 13.85/16.53/16.96 & 24.43/43.76/43.70 & 49.36/79.41/79.04 & 6.24/14.73/14.68 & 5.89/13.82/13.69 \\
    \rev{TrajODE} & \rev{31.10/39.07/57.79} & \rev{55.973/67.39/88.02} & \rev{5.75/14.01/20.42} & \rev{4.64/10.45/16.98} & \rev{36.74/37.31/37.57} & \rev{64.81/54.58/72.28} & \rev{9.43/8.32/15.98} & \rev{8.41/7.59/15.60} \\
    t2vec & 54.47/55.67/55.63 & 86.80/87.71/88.16 & 24.41/24.96/24.25 & 22.84/23.29/22.15 & 37.81/38.14/38.71 & 72.82/73.76/74.35 & 10.27/10.70/10.83 & 9.44/9.29/9.77 \\
    Robust DAA & 5.84/8.13/5.84 & 15.74/20.10/16.59 & 0.11/0.50/0.11 & 0.01/0.42/0.01 & 4.88/5.54/5.76 & 14.27/14.81/16.77 & 0.21/0.31/0.31 & 0.04/0.22/0.15 \\
    TrajectorySim & 30.42/34.22/33.48 & 57.10/60.20/60.32 & 8.33/10.93/9.30 & 8.29/11.29/9.51 & 19.42/22.90/22.36 & 43.07/49.23/48.23 & 2.60/3.49/3.25 & 2.22/3.02/2.77 \\
    PreCLN & 12.48/53.06/48.45 & 31.92/87.63/83.90 & 2.26/20.33/14.55 & 2.09/18.03/11.58 & 9.58/37.35/36.50 & 24.79/73.14/72.00 & 0.62/9.96/8.99 & 0.34/8.66/7.46 \\
    \rev{TrajCL} & \rev{48.73/53.23/53.65} & \rev{82.40/87.50/87.79} & \rev{18.49/20.28/20.91} & \rev{16.91/17.97/17.97} & \rev{29.59/37.48/37.87} & \rev{58.07/71.78/72.29} & \rev{6.85/9.23/9.41} & \rev{5.52/7.85/8.06} \\
    START & 47.21/55.14/55.28 & 79.72/87.83/86.53 & 18.31/21.46/23.81 & 17.13/19.14/21.84 & 28.92/36.13/36.87 & 59.12/71.33/72.96 & 6.35/9.17/9.96 & 5.63/7.63/8.41 \\
    LightPath & 51.00/50.17/55.99 & 84.11/84.01/88.65 & 18.62/17.91/23.04 & 16.72/17.03/21.23 & 31.48/28.19/33.51 & 63.55/58.52/66.94 & 7.16/5.09/8.06 & 6.36/3.67/7.20 \\
    \rev{MMTEC} & \rev{38.44/58.38/58.04} & \rev{68.52/90.20/89.97} & \rev{7.10/23.76/23.13} & \rev{5.90/21.26/20.44} & \rev{14.98/37.69/37.86} & \rev{34.24/72.51/72.48} & \rev{1.68/9.13/9.21} & \rev{1.36/8.00/7.84} \\
    \bottomrule
    \end{tabular}
}
\end{threeparttable}
\label{table:destination-prediction}
\end{table*}

\begin{table*}
\centering
\caption{\rev{Performance comparison of different approaches on the arrival time estimation task.}}
\begin{threeparttable}
    \begin{tabular}{c|ccc|ccc}
    \toprule
    Strategy & \multicolumn{6}{c}{wo finetune / wo pretrain / full} \\
    \midrule
    \multirow{2}{*}{Method} & \multicolumn{3}{c|}{Chengdu} & \multicolumn{3}{c}{Porto} \\
    \cline{2-7}
    & MAE(minutes) & RMSE(minutes) & MAPE(\%) & MAE(minutes) & RMSE(minutes) & MAPE(\%) \\
    \midrule
    \rev{FVTI} & \rev{1.084/1.962/0.810} & \rev{1.944/2.742/1.209} & \rev{27.73/33.84/12.15} & \rev{2.018/2.320/1.673} & \rev{1.968/1.968/2.566} & \rev{30.27/30.27/29.01} \\
    GCM & 0.513/1.988/0.445 & 0.717/3.332/0.641 & 8.16/7.11/7.12 & 1.540/1.429/1.447 & 2.421/2.236/2.299 & 24.67/23.64/23.09 \\
    \rev{POI2Vec} & \rev{1.020/1.015/1.010} & \rev{1.390/1.375/1.380} & \rev{12.12/11.98/12.05} & \rev{2.020/2.030/2.010} & \rev{2.790/2.780/2.770} & \rev{29.80/29.50/29.60} \\
    \rev{TALE} & \rev{0.970/0.971/0.975} & \rev{1.270/1.270/1.265} & \rev{10.85/10.85/10.85} & \rev{1.925/1.935/1.950} & \rev{2.650/2.670/2.660} & \rev{27.50/27.65/27.60} \\
    \rev{CTLE} & \rev{0.941/0.974/0.799} & \rev{1.343/1.251/1.251} & \rev{11.09/11.52/11.52} & \rev{0.923/0.876/0.876} & \rev{1.962/2.036/2.036} & \rev{12.23/12.95/12.95} \\
    \rev{Toast} & \rev{0.892/0.914/0.769} & \rev{1.220/1.296/1.152} & \rev{10.98/11.49/11.49} & \rev{1.245/1.368/1.171} & \rev{2.581/2.874/2.278} & \rev{12.93/13.36/22.63} \\
    DTC & 1.732/0.604/0.478 & 1.848/0.906/0.632 & 29.71/9.25/7.63 & 3.211/0.899/0.940 & 3.840/1.613/1.657 & 64.55/12.53/13.63 \\
    trajectory2vec & 0.093/0.303/0.047 & 0.101/0.342/0.065 & 1.50/5.41/0.77 & 0.406/0.064/0.126 & 0.553/0.110/0.162 & 6.01/1.08/2.34 \\
    TremBR & 0.147/0.027/0.017 & 0.245/0.041/0.023 & 2.35/0.43/0.27 & 0.970/0.034/0.028 & 1.726/0.053/0.041 & 13.12/0.51/0.43 \\
    CAETSC & 0.423/0.330/0.314 & 0.581/0.470/0.452 & 6.82/5.31/5.04 & 1.337/0.997/1.164 & 2.145/1.647/1.927 & 18.91/14.51/16.69 \\
    GM-VSAE & 0.225/0.053/0.046 & 0.316/0.067/0.059 & 3.71/0.82/0.78 & 1.089/0.169/0.131 & 1.888/0.225/0.226 & 15.02/2.92/2.17 \\
    \rev{TrajODE} & \rev{0.774/1.821/0.922} & \rev{1.054/1.902/1.100} & \rev{30.96/30.00/16.14} & \rev{2.733/0.855/0.991} & \rev{3.961/1.931/0.688} & \rev{52.08/13.52/10.55} \\
    t2vec & 0.361/0.223/0.221 & 0.513/0.317/0.321 & 5.77/3.65/3.60 & 1.227/0.895/0.941 & 2.060/1.606/1.626 & 17.77/12.69/13.58 \\
    Robust DAA & 4.420/0.851/1.560 & 4.841/1.256/1.843 & 75.78/12.89/26.93 & 2.782/2.730/2.542 & 3.782/3.819/3.443 & 50.11/41.66/46.52 \\
    TrajectorySim & 0.164/0.083/0.075 & 0.261/0.112/0.096 & 2.56/1.36/1.25 & 0.664/0.103/0.145 & 1.129/0.154/0.239 & 9.44/1.71/2.35 \\
    PreCLN & 0.329/0.212/0.197 & 0.430/0.286/0.314 & 5.70/2.95/2.99 & 0.749/0.295/0.408 & 1.440/0.435/0.641 & 10.15/4.58/6.34 \\
    \rev{TrajCL} & \rev{0.263/0.173/0.133} & \rev{0.315/0.221/0.216} & \rev{5.55/2.94/2.21} & \rev{0.925/0.589/0.698} & \rev{1.980/1.857/1.921} & \rev{13.55/9.49/9.13} \\
    START & 0.306/0.331/0.063 & 0.427/0.531/0.084 & 4.84/5.03/1.06 & 0.440/0.462/0.079 & 0.614/0.622/0.121 & 8.36/7.96/1.45 \\
    LightPath & 0.284/0.310/0.142 & 0.391/0.390/0.170 & 4.63/5.30/2.46 & 0.414/0.780/0.162 & 0.574/0.971/0.216 & 7.18/14.44/3.06 \\
    \rev{MMTEC} & \rev{0.316/0.239/0.215} & \rev{0.202/0.350/0.327} & \rev{3.22/3.97/3.47} & \rev{0.577/0.809/0.713} & \rev{1.732/2.521/2.077} & \rev{8.06/10.02/9.59} \\
    \bottomrule
    \end{tabular}
\end{threeparttable}

\label{table:arrival-time}
\end{table*}

\begin{table*}
\centering
\caption{\rev{Performance comparison of different approaches on the trajectory classification task.}}
\begin{threeparttable}
    \resizebox{1.0\linewidth}{!}{
    \begin{tabular}{c|cccc|cccc}
    \toprule
    Strategy & \multicolumn{8}{c}{wo finetune / wo pretrain / full} \\
    \midrule
    \multirow{2}{*}{Method} &\multicolumn{4}{c|}{Chengdu} & \multicolumn{4}{c}{Porto} \\ 
    \cline{2-9}
    & Acc@1(\%) & Acc@5(\%) & Recall(\%) & f1(\%) & 
    Acc@1(\%) & Acc@5(\%) & Recall(\%) & f1(\%) \\
    \midrule
    
    \rev{FVTI} & \rev{1.01/1.31/1.34} & \rev{4.52/5.78/6.15} & \rev{1.17/1.28/1.01} & \rev{0.93/0.59/0.54} & \rev{1.09/1.16/1.37} & \rev{6.55/6.05/6.29} & \rev{0.91/0.92/1.01} & \rev{0.70/0.53/0.64} \\
    GCM & 9.53/11.09/10.15 & 11.14/12.03/11.67 & 7.46/8.57/8.15 & 6.87/7.90/7.53 & 5.07/4.94/5.18 & 14.63/14.71/15.33 & 4.53/4.30/4.57 & 3.39/3.17/3.22 \\
    \rev{POI2Vec} & \rev{5.80/5.65/7.10} & \rev{7.52/6.99/8.80} & \rev{4.91/4.84/5.82} & \rev{4.34/4.26/5.19} & \rev{3.74/3.68/4.56} & \rev{11.89/11.87/13.22} & \rev{3.04/2.88/3.51} & \rev{2.15/2.08/2.55} \\
    \rev{TALE} & \rev{5.95/6.75/7.55} & \rev{7.43/7.79/8.50} & \rev{4.88/5.92/5.65} & \rev{4.12/5.34/4.95} & \rev{3.51/4.32/4.13} & \rev{11.31/12.81/12.61} & \rev{2.98/3.25/3.14} & \rev{2.12/2.44/2.38} \\
    \rev{CTLE} & \rev{1.71/1.42/1.71} & \rev{5.67/5.91/5.57} & \rev{1.25/2.04/1.64} & \rev{0.42/0.82/1.33} & \rev{1.50/1.34/1.63} & \rev{6.73/6.61/6.86} & \rev{0.97/1.16/1.54} & \rev{0.06/0.77/1.21} \\
    \rev{Toast} & \rev{1.63/1.11/1.13} & \rev{5.51/5.38/4.52} & \rev{1.03/1.71/1.26} & \rev{0.38/1.00/0.96} & \rev{1.50/1.44/1.56} & \rev{6.73/6.36/6.77} & \rev{0.97/0.99/1.22} & \rev{0.06/0.48/0.93} \\
    DTC & 0.03/0.21/0.03 & 0.15/0.53/0.15 & 0.01/0.18/0.01 & 0.00/0.13/0.00 & 0.48/5.42/4.62 & 2.14/16.66/15.53 & 0.24/4.65/3.82 & 0.00/3.08/2.31 \\
    trajectory2vec & 1.22/0.03/0.61 & 3.04/0.17/1.44 & 1.06/0.02/0.48 & 0.91/0.00/0.42 & 0.92/1.04/1.16 & 3.32/4.34/4.01 & 0.43/0.79/0.62 & 0.09/0.35/0.13 \\
    TremBR & 11.17/11.33/11.55 & 12.08/12.05/12.22 & 8.76/8.34/8.89 & 8.10/7.59/8.16 & 4.89/6.04/5.65 & 14.58/17.85/16.86 & 4.30/5.25/4.92 & 2.79/3.73/3.65 \\
    CAETSC & 0.55/0.04/0.03 & 1.43/0.17/0.15 & 0.50/0.02/0.02 & 0.40/0.00/0.00 & 1.31/3.48/2.77 & 4.75/11.70/9.81 & 0.85/2.83/2.03 & 0.24/1.65/1.06 \\
    GM-VSAE & 11.96/11.99/12.17 & 12.32/12.21/12.33 & 9.84/9.05/9.65 & 9.10/8.21/8.79 & 4.88/5.69/5.72 & 15.55/16.84/17.32 & 4.25/5.08/4.98 & 2.59/3.69/3.54 \\
    \rev{TrajODE} & \rev{1.25/0.76/1.26} & \rev{6.78/5.03/6.03} & \rev{1.98/0.96/1.19} & \rev{0.41/0.17/0.30} & \rev{1.43/1.66/1.58} & \rev{6.06/7.08/6.64} & \rev{1.01/1.04/1.08} & \rev{0.34/0.15/0.23} \\
    t2vec & 9.67/11.97/11.71 & 11.42/12.25/12.23 & 7.18/9.45/8.85 & 6.48/8.65/8.11 & 5.39/5.36/5.74 & 16.47/16.55/16.91 & 4.71/4.70/5.08 & 3.43/3.23/3.75 \\
    Robust DAA & 0.03/0.28/0.04 & 0.13/0.54/0.16 & 0.01/0.26/0.01 & 0.00/0.27/0.00 & 0.84/0.67/0.98 & 3.15/2.73/3.55 & 0.37/0.35/0.46 & 0.02/0.10/0.07 \\
    TrajectorySim & 8.60/9.66/9.70 & 10.65/11.10/11.24 & 6.64/7.42/7.45 & 6.34/7.12/7.13 & 2.55/2.79/3.02 & 9.20/10.39/10.54 & 2.07/2.31/2.51 & 1.31/1.34/1.65 \\
    PreCLN & 11.54/6.65/6.69 & 12.36/7.97/8.12 & 9.49/5.48/5.97 & 8.81/5.36/5.67 & 4.48/2.06/1.96 & 12.39/4.97/5.37 & 4.02/1.04/1.17 & 3.20/0.95/1.22 \\
    \rev{TrajCL} & \rev{1.66/2.01/1.42} & \rev{6.58/5.78/5.78} & \rev{1.51/1.55/1.29} & \rev{1.08/0.92/0.88} & \rev{1.30/1.32/1.34} & \rev{5.56/6.63/6.62} & \rev{1.07/0.92/0.99} & \rev{0.63/0.37/0.62} \\
    START & 11.63/7.65/7.09 & 12.42/9.95/10.13 & 9.59/6.51/6.28 & 8.97/6.07/5.92 & 4.52/6.18/6.34 & 13.66/19.63/19.88 & 4.13/5.46/5.61 & 3.12/3.80/3.79 \\
    LightPath & 2.18/0.03/0.97 & 4.24/0.17/2.19 & 1.89/0.01/0.91 & 1.74/0.00/0.78 & 4.76/1.04/0.98 & 14.40/3.81/3.56 & 4.34/0.57/0.54 & 2.91/0.08/0.07 \\
    \rev{MMTEC} & \rev{8.27/7.19/7.27} & \rev{12.94/9.18/9.52} & \rev{8.06/6.13/6.45} & \rev{6.23/5.41/6.01} & \rev{3.95/1.63/1.34} & \rev{7.22/6.56/6.62} & \rev{1.97/1.15/1.00} & \rev{1.33/0.84/0.38} \\
    
    \bottomrule
    \end{tabular}
    }
\end{threeparttable}

\label{table:classification}
\end{table*}

\subsection{Building Existing Methods with UniTE}
UniTE is designed to modularize the implementation of existing and new methods for the pre-training of trajectory embeddings. 
\rev{Table~\ref{tab:methods} presents a summary of the UniTE modules needed to build existing methods introduced in Section~\ref{sec:survey}. The table is organized by the methods' pre-training frameworks, corresponding to the classification in Figure~\ref{fig:methods}.}

\section{Experiments}
\label{sec:experiments}
\rev{To illustrate the operation of UniTE pipeline and also to provide guidance on the effectiveness and characteristics of different modules,} we implement existing methods with UniTE and perform experiments to evaluate their effectiveness using the datasets and downstream adapters included in UniTE.

\subsection{Settings}
We conduct experiments on the Chengdu and Porto datasets, described in Table~\ref{table:datasets}. For consistency, both datasets are standardized to a sampling interval of 15 seconds. Trajectories with fewer than six points are excluded. We sort the trajectories by their start time and split the datasets into training, evaluation, and testing sets in an 8:1:1 ratio. Both pre-training and fine-tuning are performed on the training set. The fine-tuning process includes an early-stopping mechanism with 10 epochs of tolerance, based on metrics calculated on the evaluation set. The final metrics are calculated on the testing set.

We evaluate all existing trajectory embedding methods introduced in Section~\ref{sec:survey}, implementing them with the UniTE pipeline. In terms of downstream adaptors discussed in Section~\ref{sec:downstream-adaptor}, we use destination prediction, arrival time estimation, and trajectory classification adapters. These cover downstream tasks in three representative scenarios: incomplete trajectory sequence, incomplete trajectory features, and complete trajectory sequence. All the three types of training strategies covered in Section~\ref{sec:training-strategy} are applied in the experiments.

\subsection{Overall Performance}
Tables~\ref{table:destination-prediction} to~\ref{table:classification} compare the overall performance of selected methods at destination prediction, arrival-time estimation, and trajectory classification.
\rev{Each table cell reports one metric of each method under the three training strategies, with / separating different strategies. From the results, we can draw the following insights.}

\begin{figure}[htbp]
    \setlength{\abovecaptionskip}{-8pt}
    \centering
    \input{plot/pretrainer-comparison}
    \caption{\rev{Performance comparison of methods under the \textit{wo finetune} strategy. Methods are grouped by their pre-trainers.}}
    \label{fig:pretrainer-comparison}
\end{figure}

\textbf{Comparison within the same training strategy} provides insights into the effectiveness of the learnable components and pre-trainers used. The \textit{wo finetune} strategy, which fixes the trainable parameters after pre-training, highlights the effectiveness of the chosen pre-trainer. \rev{A visual comparison is provided in Figure~\ref{fig:pretrainer-comparison}, in which methods are arranged into three groups corresponding to the pre-trainers they use: contrastive (grey), generative (red), and hybrid (yellow).} We present the following observations:
\begin{enumerate}
    \item The contrastive pre-trainer, which focuses on contrasting different views of one or multiple trajectories, enhances the performance of trajectory embeddings on tasks that rely on the global aspect of a full trajectory, such as trajectory classification.
    \item The generative pre-trainer, which targets the reconstruction of the spatiotemporal features of a trajectory, improves the performance of trajectory embeddings on tasks that rely on local spatiotemporal correlations, such as trajectory prediction.
    \item The hybrid pre-trainer, combining the benefits of both generative and contrastive pre-training, enables trajectory embeddings to perform competitively across multiple types of tasks, as demonstrated by START and LightPath.
\end{enumerate}
Discrepancies between different types of pre-trainers are also discussed in other studies on the pre-training of embeddings~\cite{liu2022self,lin2023pre}.

\begin{figure}[htbp]
    \setlength{\abovecaptionskip}{-8pt}
    \centering
    \input{plot/encoder-comparison}
    \caption{\rev{Performance comparison of methods under the \textit{wo pretrain} strategy. Methods are grouped by their encoders.}}
    \label{fig:encoder-comparison}
\end{figure}

The \textit{wo pretrain} strategy learns a method's parameters end-to-end for a specific task, with the performance of embeddings influenced primarily by the preprocessor and learnable components. \rev{A visual comparison is provided in Figure~\ref{fig:encoder-comparison}, in which methods are arranged into three groups corresponding to the encoders they use: RNN-based (grey), Transformer-based (red), and CNN-based (yellow).} We have the following observations:
\begin{enumerate}
    \item Methods using RNN- or Transformer-based encoders generally outperform those using CNN-based encoders at destination prediction, as this task emphasizes modeling sequential correlations and accurately capturing spatial information of trajectories.
    \item Methods that emphasize time features exhibit superior performance at the arrival time estimation that benefits from temporal information.
\end{enumerate}

\textbf{Comparison between different training strategies} evaluates the effectiveness of the pre-training and fine-tuning processes. \rev{The radar charts in Figure~\ref{fig:radar-destination} and~\ref{fig:radar-classification} provide a visual comparison of different training strategies for each method, in which the axis \textit{wop.}, \textit{wof.} and \textit{full} correspond to the \textit{wo finetune}, \textit{wo pretrain}, and \textit{full} strategies, respectively.} We observe the following:
\begin{enumerate}
    \item The \textit{full} strategy, which involves pre-training followed by fine-tuning, aligns with most methods for the pre-training of trajectory embeddings and yields optimal performance in most cases. Pre-training helps methods gain a universal understanding of trajectories, while the fine-tuning process further adjusts the methods to fit the specific task better.
    \item PreCLN, START, and LightPath perform better using the \textit{wo finetune} strategy at trajectory classification compared to the other two strategies. This may be due to the task-specific labels being of low quality, such as the uneven distribution of class labels in trajectory classification. This observation highlights one of the benefits of pre-training, which is to enhance performance when task-specific labels are insufficient to support effective end-to-end training.
\end{enumerate}

\begin{figure*}
    \setlength{\abovecaptionskip}{-10pt}
    \centering
    \input{plot/radar-destination}
    \caption{\rev{Destination prediction (Acc@1, \%) performance of different methods and training strategies on Chengdu.}}
    \label{fig:radar-destination}
\end{figure*}

\begin{figure*}
    \setlength{\abovecaptionskip}{-10pt}
    \centering
    \input{plot/radar-classification}
    \caption{\rev{Trajectory classification (Acc@1, \%) performance of different methods and training strategies on Porto.}}
    \label{fig:radar-classification}
\end{figure*}

\subsection{Efficiency}
Table~\ref{table:efficiency} reports on the efficiency metrics for the comparison methods.
\rev{Each table cell reports the one metric of each method on the two datasets, with / separating different datasets.}

The model size and embedding time are affected primarily by the complexity of the preprocessors and learnable components within a method. Additionally, the pre-training time depends on the specific pre-trainer used by a method. We observe that methods combining tokenization or map-matching preprocessors with index-fetching feature embedders tend to have larger model sizes, as each token or road segment is assigned an embedding vector. Methods employing Transformer-based encoders generally exhibit higher pre-training and embedding times compared to those using RNN-based encoders, due to the higher computational costs of Transformers.
Some methods, such as t2vec and TrajectorySim, show relatively high pre-training times, primarily because their preprocessors take longer time to run. 
\rev{TrajODE also takes a long time to pre-train due to its ODE-based encoder and decoder, which is computationally expensive.}
Overall, understanding these efficiency metrics is crucial for selecting an appropriate method based on the available computational resources, the requirements of the task, and the preferred balance between accuracy and efficiency.

\begin{table}
\centering
\caption{Efficiency comparison of different approaches.}
\begin{threeparttable}
    \begin{tabular}{c|ccc}
    \toprule
    Dataset & \multicolumn{3}{c}{Chengdu / Porto} \\
    \midrule
    \multirow{2}{*}{Method} & Model size & Pretrain time & Embed time \\
    & (MBytes) & (min/epoch) & (sec) \\
    \midrule
    \rev{FVTI} & \rev{4.883 / 4.883} & \rev{1.488 / 1.917} & \rev{0.504 / 0.562} \\
    GCM & 1.774 / 1.989 & 1.427 / 2.527 & 1.174 / 1.214 \\
    \rev{POI2Vec} & \rev{3.433 / 2.301} & \rev{1.534 / 1.936} & \rev{1.217 / 1.309} \\
    \rev{TALE} & \rev{5.127 / 3.892} & \rev{1.612 / 1.950} & \rev{1.267 / 1.387} \\
    \rev{CTLE} & \rev{8.045 / 7.357} & \rev{2.253 / 2.653} & \rev{1.264 / 3.116} \\
    \rev{Toast} & \rev{8.045 / 7.357} & \rev{3.065 / 2.732} & \rev{1.283 / 3.181} \\
    DTC & 27.239 / 21.057 & 4.354 / 3.492 & 6.714 / 8.836 \\
    trajectory2vec & 0.535 / 0.535 & 4.683 / 4.416 & 1.407 / 1.606 \\
    TremBR & 12.640 / 11.820 & 0.790 / 0.954 & 1.064 / 1.179 \\
    CAETSC & 0.386 / 0.386 & 0.588 / 1.138 & 4.605 / 4.268 \\
    GM-VSAE & 5.467 / 2.629 & 0.682 / 1.972 & 1.453 / 0.986 \\
    \rev{TrajODE} & \rev{0.987 / 0.987} & \rev{364.208 / 551.192} & \rev{0.496 / 0.848}\\
    t2vec & 15.341 / 15.044 & \rev{1.120 / 4.316} & 5.994 / 5.850 \\
    Robust DAA & 0.022 / 0.022 & 0.217 / 0.201 & 3.525 / 3.617 \\
    TrajectorySim & 12.110 / 12.212 & \rev{1.688 / 5.030} & 6.184 / 7.016 \\
    PreCLN & 56.625 / 55.048 & 7.988 / 12.523 & 5.251 / 6.875 \\
    \rev{TrajCL} & \rev{8.249 / 7.975} & \rev{1.710 / 2.013} & \rev{9.080 / 9.850} \\
    START & 25.527 / 24.159 & 18.357 / 34.806 & 5.649 / 7.218 \\
    LightPath & 12.074 / 11.938 & 2.791 / 2.738 & 9.479 / 8.932 \\
    \rev{MMTEC} & \rev{2.958 / 2.821} & \rev{0.438 / 0.513} & \rev{3.174 / 3.528} \\
    \bottomrule 
    \end{tabular}
\end{threeparttable}
\label{table:efficiency}
\end{table}



\section{Conclusion}\label{sec:conclusion}
We present UniTE, a comprehensive survey and a unified pipeline aimed at accelerating advances in methods for the pre-training of trajectory embeddings. 
The survey compiles an extensive list of existing methods, including those explicitly targeting universal trajectory embeddings and those that implicitly employ pre-training techniques tailored for specific tasks. 
The unified pipeline standardizes the implementation and evaluation of methods for the pre-training methods, facilitating the reproduction of existing methods and the development of new ones. 
Together, the survey and pipeline offer a thorough academic and technical resource, which we hope will accelerate research in this field.

\section*{Acknowledgments}
This work was supported by the National Natural Science Foundation of China (No. 62272033).

\bibliographystyle{IEEEtran}
\bibliography{reference}

\begin{IEEEbiography}[{\includegraphics[width=1in,height=1.25in,clip,keepaspectratio]{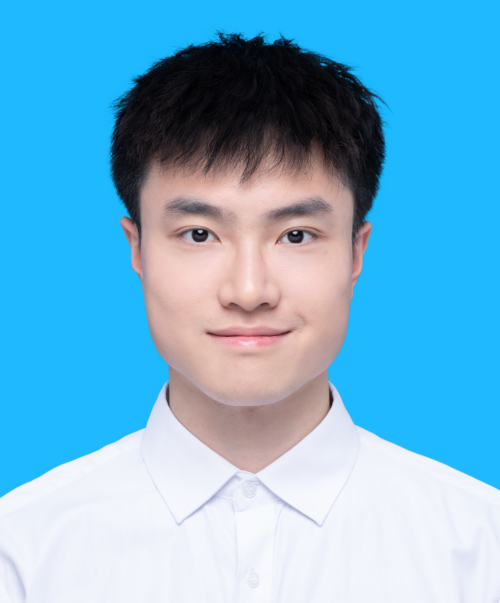}}]{Yan Lin} received the B.S. degree in computer science from Beijing Jiaotong University, Beijing, China, in 2019.

He is currently working toward the Ph.D. degree in the School of Computer and Information Technology, Beijing Jiaotong University. His research interests include spatiotemporal data mining and representation learning.
\end{IEEEbiography}


\begin{IEEEbiography}
[{\includegraphics[width=1in,height=1.25in,clip,keepaspectratio]{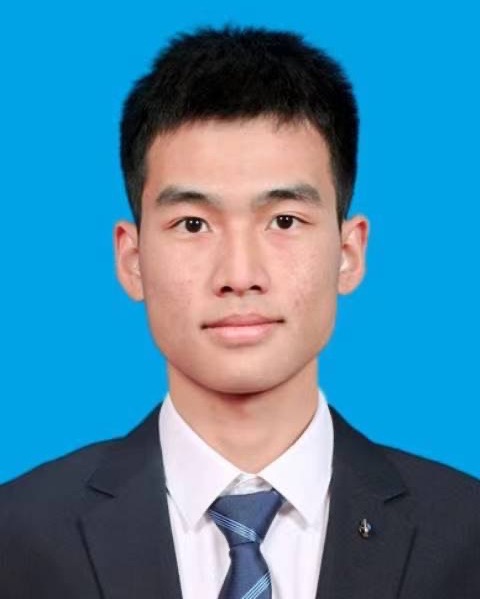}}]
{Zeyu Zhou} received the B.S. degree in mathematics and applied mathematics from Beijing Jiaotong University, Beijing, China, in 2022.

He is currently working toward the M.S. degree in the School of Computer and Information technology, Beijing Jiaotong University. His research interests focus on deep learning and data mining, particularly their applications in spatiotemporal data mining.
\end{IEEEbiography}


\begin{IEEEbiography}
[{\includegraphics[width=1in,height=1.25in,clip,keepaspectratio]{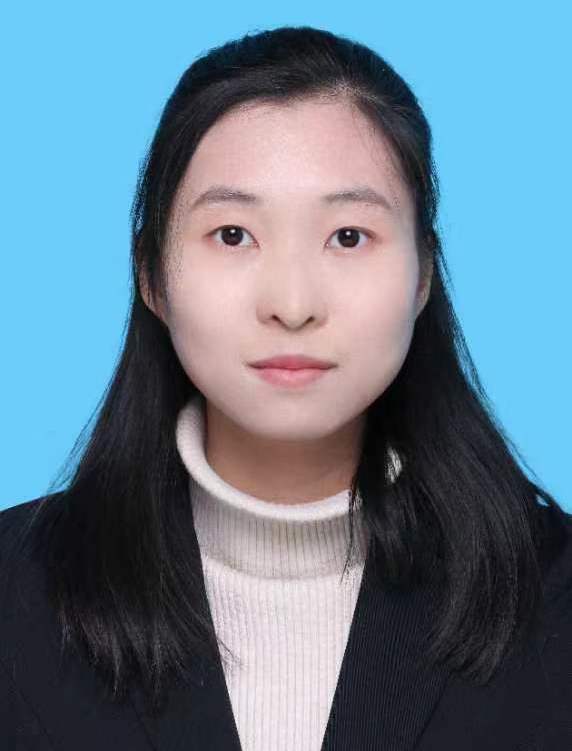}}]
{Yichen Liu} received the B.S. degree in computer science from Beijing Jiaotong University, Beijing, China, in 2023.

She is currently working toward the M.S. degree in the School of Computer and Information technology, Beijing Jiaotong University. Her research interests focus on deep learning and data mining, especially their applications in spatiotemporal data mining.
\end{IEEEbiography}


\begin{IEEEbiography}
[{\includegraphics[width=1in,height=1.25in,clip,keepaspectratio]{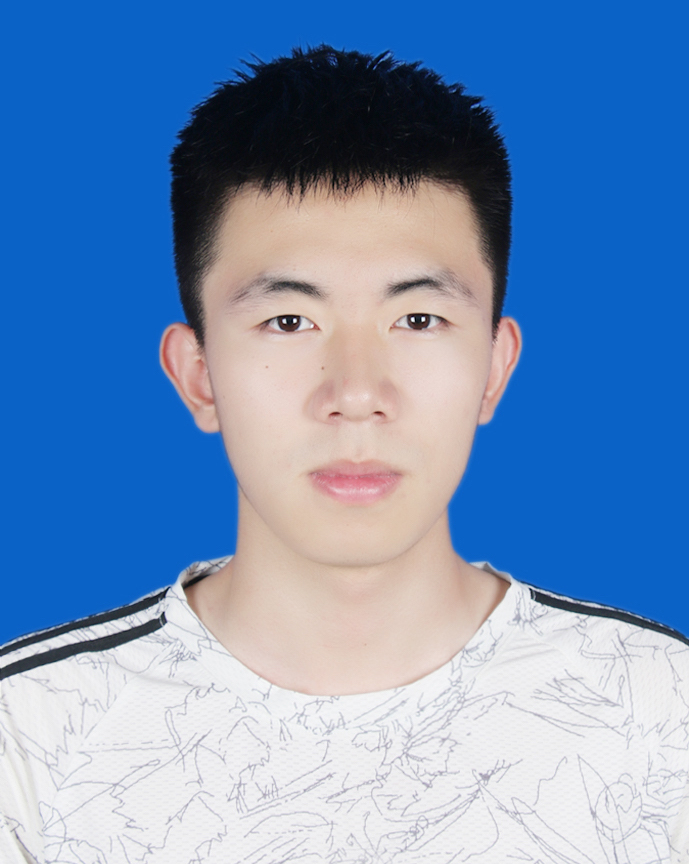}}]
{Haochen Lv} received the B.S. degree in computer science from Beijing Jiaotong University, Beijing, China, in 2024. He is currently working toward the Ph.D. degree in the School of Computer and Information Technology, Beijing Jiaotong University. 

His research interests focus on spatial-tempora data mining and spatiotemporal graph.
\end{IEEEbiography}


\begin{IEEEbiography}[{\includegraphics[width=1in,height=1.25in,clip,keepaspectratio]{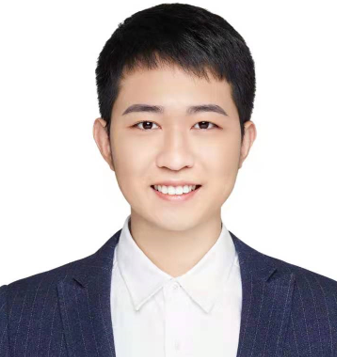}}]{Haomin Wen} received the B.S. degree in computer science and technology
from Beijing Jiaotong University, Beijing, China, in 2019, where he is currently pursuing the Ph.D. degree in computer science with the School of Computer and Information Technology.

His current research interests include spatial-temporal data mining and intelligent transportation technology.
\end{IEEEbiography}


\begin{IEEEbiography}[{\includegraphics[width=1in,height=1.25in,clip,keepaspectratio]{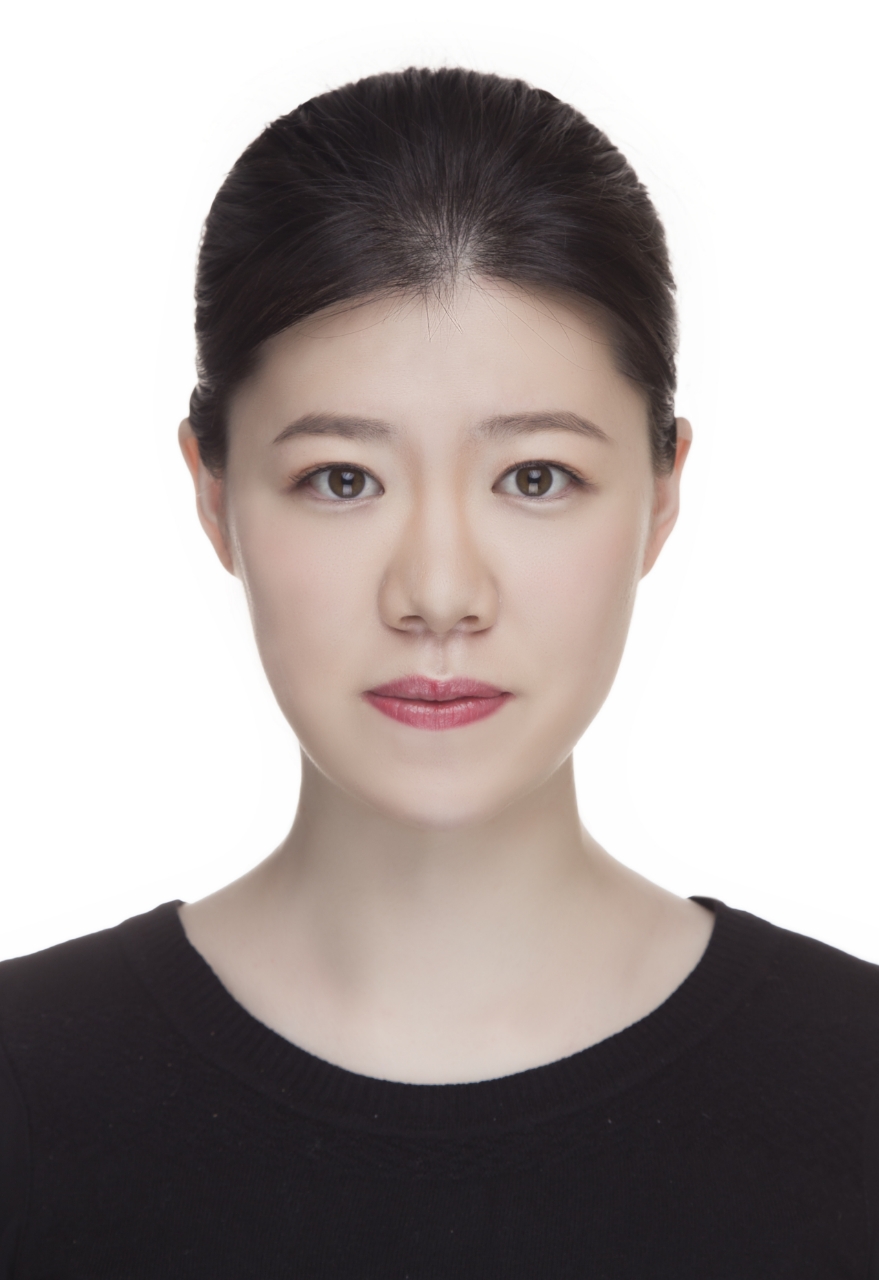}}]{Tianyi Li} received the Ph.D. degree from Aalborg University, Denmark, in 2022. 

She is an assistant professor at the Department of Computer Science, Aalborg University. Her research concerns primarily data management and analytics, intelligent transportation, machine learning, and database technology. 
\end{IEEEbiography}


\begin{IEEEbiography}[{\includegraphics[width=1in,height=1.25in,clip,keepaspectratio]{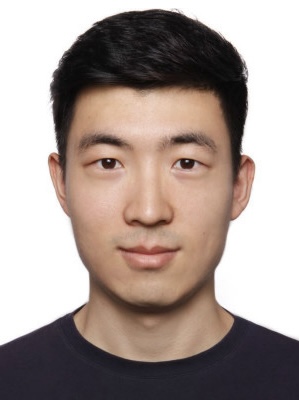}}]{Yushuai Li} received the Ph.D. degree in control theory and control engineering from Northeastern University, Shenyang, China, in 2019. 

He is currently an assistant professor at the Department of Computer Science, Aalborg University. His research interests include machine learning, digital twin, digital energy, and intelligent transportation systems.
\end{IEEEbiography}


\begin{IEEEbiography}[{\includegraphics[width=1in,height=1.25in,clip,keepaspectratio]{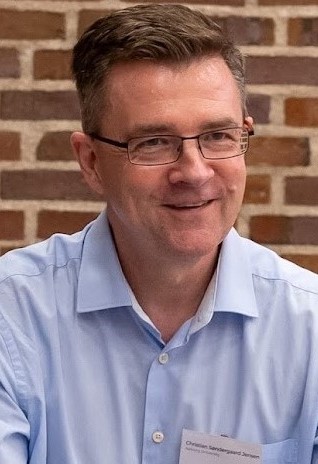}}]{Christian S. Jensen} received the Ph.D. degree from Aalborg University in 1991 after 2 1/2 years of study at University of Maryland, and he received the Dr.Techn. degree from Aalborg University in 2000.

He is a Professor at the Department of Computer Science, Aalborg University. 
His research concerns primarily temporal and spatiotemporal data management and analytics, including indexing and query processing, data mining, and machine learning.
\end{IEEEbiography}


\begin{IEEEbiography}[{\includegraphics[width=1in,height=1.25in,clip,keepaspectratio]{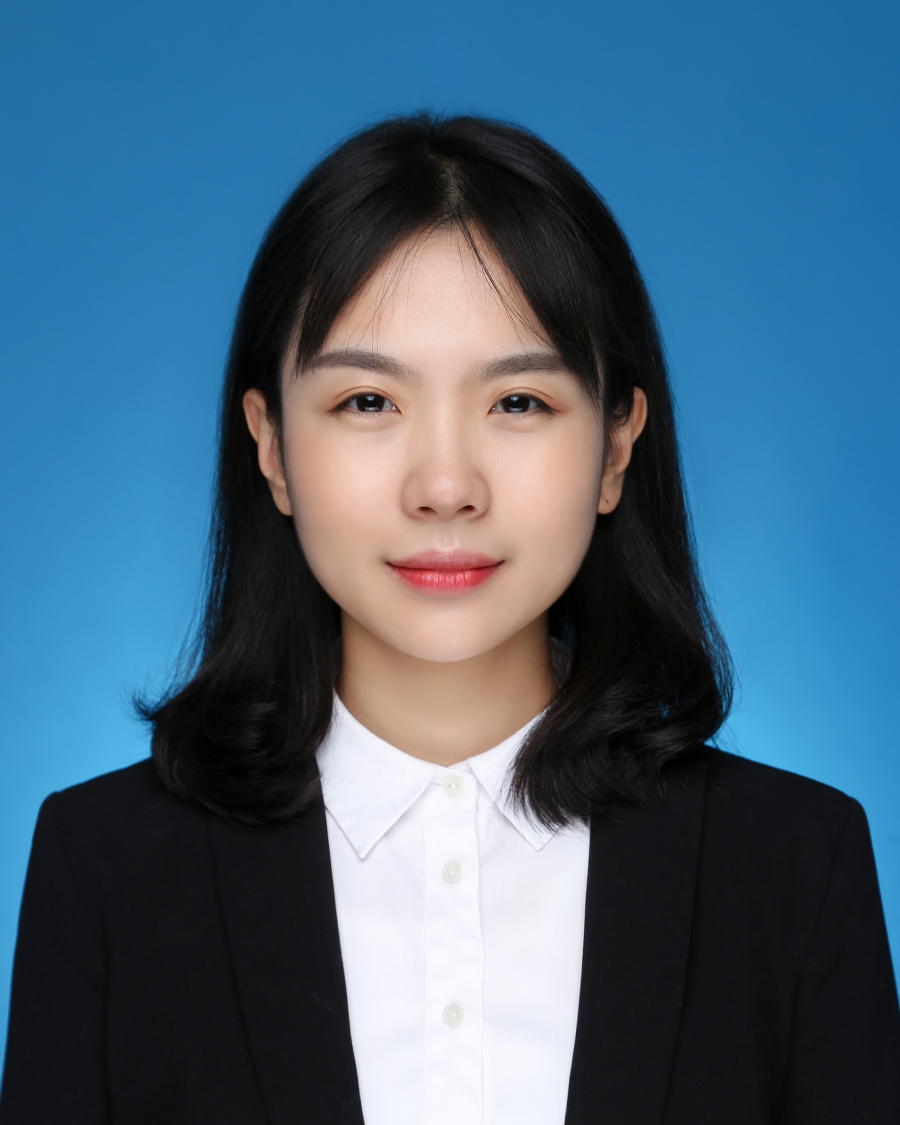}}]{Shengnan Guo} received the Ph.D. degree in computer science from Beijing Jiaotong University, Beijing, China, in 2021.

She is an associate professor at the School of Computer and Information Technology, Beijing Jiaotong University. Her research interests focus on spatial-temporal data mining and intelligent transportation systems.
\end{IEEEbiography}


\begin{IEEEbiography}[{\includegraphics[width=1in,height=1.25in,clip,keepaspectratio]{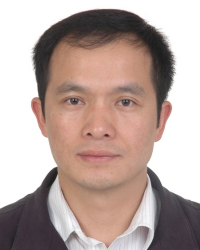}}]{Youfang Lin} received the Ph.D. degree in signal and information processing  from Beijing Jiaotong University, Beijing, China, in 2003.

He is a professor with the School of Computer and Information Technology, Beijing Jiaotong University. His main fields of expertise and current research interests include big data technology, intelligent systems, complex networks, and traffic data mining.
\end{IEEEbiography}


\begin{IEEEbiography}[{\includegraphics[width=1in,height=1.25in,clip,keepaspectratio]{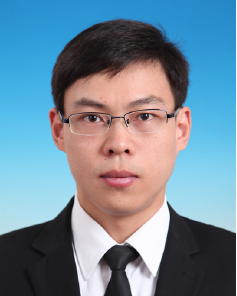}}]{Huaiyu Wan} received the Ph.D. degree in computer science and technology from Beijing Jiaotong University, Beijing, China, in 2012.

He is a professor with the School of Computer and Information Technology, Beijing Jiaotong University. His current research interests focus on spatiotemporal data mining, social network mining, information extraction, and knowledge graph.
\end{IEEEbiography}

\end{document}